\documentclass[preprint,12pt,3p]{elsarticle}
\usepackage{tabularx}
\usepackage{caption}
\usepackage{geometry}
\usepackage{pifont} 
\usepackage{booktabs} 
\usepackage{amsmath}
\usepackage{booktabs}
\usepackage{makecell}
\usepackage{url}
\usepackage{graphicx}
\usepackage{balance}
\usepackage{amsmath}
\usepackage{subfigure}
\usepackage{graphicx}
\usepackage{balance}
\usepackage{comment}
\usepackage{array}
\usepackage{lipsum}
\usepackage{caption}
\usepackage{amsmath}
\usepackage{array}
\usepackage{geometry}
\usepackage{pifont} 
\usepackage{booktabs} 
\newcommand{\cmark}{\checkmark}  
\newcommand{\xmark}{\times} 
\usepackage{graphicx}   
  
\usepackage{multirow}   
\usepackage{booktabs}   
\usepackage{amssymb}    
\usepackage{caption}   
\usepackage{amsmath}        
\usepackage{blindtext}
\usepackage{enumitem}
\usepackage{lipsum}
\usepackage{capt-of}
\usepackage{tikz}
\usetikzlibrary{arrows,positioning,automata}
\usepackage{balance}
\usepackage{amsmath}
\usepackage[T1]{fontenc}
\usepackage{algorithm}
\usepackage{algpseudocode}
\usepackage{subfigure}
\usepackage{graphicx}
\usepackage[utf8]{inputenc}
\usepackage{balance}
\usepackage{multirow}
\usepackage{multicol}
\usepackage{comment}
\usepackage{array}
\usepackage{soul}
\usepackage[justification=centering]{caption}
\usepackage[lighttt]{lmodern}
\usepackage{libertine}

\usepackage[utf8]{inputenc}
\definecolor{turquoise}{RGB}{64, 224, 208}
\definecolor{darkblue}{RGB}{0, 0, 139}
\definecolor{green}{RGB}{0, 128, 0}
\usepackage{graphicx}
\usepackage{subcaption}
\usepackage{amssymb}
\usepackage{float}
\usepackage{hyperref}
\usepackage[perpage]{footmisc}
\usepackage{hyperref} 
\usepackage{zref-user}
\definecolor{bluee}{RGB}{45, 104, 196}
\definecolor{yellow-or}{RGB}{247, 167, 12}
\journal{}
\begin{document}
\begin{frontmatter}
\title{Emotion-aware Dual Cross-Attentive Neural Network with Label Fusion for Stance Detection in Misinformative Social Media Content}

\author[label1]{Lata Pangtey}

\address[label1]{Department of Computer Science and Engineering, Indian Institute of Technology (IIT) Indore, Indore 453552, India}

\ead{ms2304101009@iiti.ac.in}
\author[label1]{Mohammad Zia Ur Rehman}
\ead{phd2101201005@iiti.ac.in}

\author[label1]{Prasad Chaudhari}
\ead{ms2204101003@iiti.ac.in}
\author[label1]{Shubhi Bansal}
\ead{phd2001201007@iiti.ac.in}
\author[label1]{Nagendra Kumar\corref{cor1}}
\ead{nagendra@iiti.ac.in}
\cortext[cor1]{Corresponding author}
\begin{abstract}
The rapid evolution of social media has generated an overwhelming volume of user-generated content, conveying implicit opinions and contributing to the spread of misinformation. The method aims to enhance the detection of stance where misinformation can polarize user opinions. Stance detection has emerged as a crucial approach to effectively analyze underlying biases in shared information and combating misinformation. This paper proposes a novel method for \textbf{S}tance \textbf{P}rediction through a \textbf{L}abel-fused dual cross-\textbf{A}ttentive \textbf{E}motion-aware neural \textbf{Net}work (SPLAENet) in misinformative social media user-generated content. The proposed method employs a dual cross-attention mechanism and a hierarchical attention network to capture inter and intra-relationships by focusing on the relevant parts of source text in the context of reply text and vice versa.
We incorporate emotions to effectively distinguish between different stance categories by leveraging the emotional alignment or divergence between the texts. 
We also employ label fusion that uses distance-metric learning to align extracted features with stance labels, improving the method's ability to accurately distinguish between stances. 
Extensive experiments demonstrate the significant improvements achieved by SPLAENet over existing state-of-the-art methods. { SPLAENet demonstrates an average gain of 8.92\% in accuracy and 17.36\% in F1-score on the RumourEval dataset. On the SemEval dataset, it achieves average gains of 7.02\% in accuracy and 10.92\% in F1-score. On the P-stance dataset, it demonstrates average gains of 10.03\% in accuracy and 11.18\% in F1-score. These results validate the effectiveness of the proposed method for stance detection in the context of misinformative social media content.}  {Our code is publicly available at:} \href{https://github.com/lata04/SPLAENet.git}{https://github.com/lata04/SPLAENet}.
\end{abstract}

\end{frontmatter}
\section{Introduction}
Social media has revolutionized the paradigm of information dissemination in the digital era, creating a dynamic landscape characterized by the vast amount of User-Generated Content (UGC) across numerous platforms. UGC includes any written content created and shared by individuals, such as blog posts, social media, comments, reviews, articles, and forum discussions. In the era of Web 2.0, the global social media user base is projected to reach approximately 5.17 billion, with individuals engaging on an average of 6.7 different social networks each month\footnote{\href{https://datareportal.com/social-media-users}{https://datareportal.com/social-media-users}}. 
While this democratization empowers voices, it also amplifies challenges like misinformation \cite{RUFFO2023100531, olan2024fake}.

Stance refers to a viewpoint on a specific topic articulated in public discourse, which may involve supporting, denying, querying, or commenting on claims, reveals biases and intentions in shared information \cite{luo2024joint}.
Stance detection uncovers underlying biases and intentions behind the shared information by analyzing the stance, facilitating the identification of sources that promote misleading narratives.
To illustrate this concept, consider the example shown in Table \ref{table:example}, which presents source-reply text pairs along with their associated stances.

\begin{table}[!ht]\normalsize
    \centering
    \caption{Examples of Source-Reply Text and the Stance}
    \begin{tabular}{|p{6.7cm}|p{6.2cm}|p{1.6cm}|}
    \hline
    \textbf{Source Text} & \textbf{Reply Text} & \textbf{Stance} \\ \hline
    Face facts: Immigrants commit fewer crimes than U.S.-born peer. & That's right by statistics. & Support \\ \hline
    What exactly is happening when you crack your joints, and is it true that it can cause arthritis? & But how do you know for sure? & Query \\ \hline
    Is it true the Earth is flat? Is there proof? Why are there people that believe it's true? & There is no proof to this and whoever says it's true is a troll or a moron. & Deny \\ \hline
    [Serious] Is it true that 85\% of people can only breathe through one nostril at a time? Who here can breathe with both nostrils? & I can breathe with both nostrils but the other one is a little weaker. The weaker one changes occasionally. & Comment \\ \hline
    \end{tabular}
    \label{table:example}
\end{table}

The source text represents the primary claim being addressed, while the reply text reflects how the responder engages with or reacts to the source text. The stance column categorizes the nature of the reply concerning the source text into four categories, i.e., Support, Query, Deny, and Comment.
In the ``Support'' stance, the source text asserts that immigrants commit fewer crimes than U.S.-born individuals. The reply, stating ``That's right by statistics,'' illustrates a supportive stance. This response affirms the original claim, indicating agreement and providing validation through references to statistics, thereby reinforcing the assertion that data support the claim about immigrants and crime rates.
The ``Query'' stance regarding cracking joints, the source text expresses concern about a potential link between cracking joints and the risk of arthritis, reflecting a negative emotional undertone related to fear or worry. The reply, however, exhibits no emotional content and instead poses a question about the certainty of this information. This contrast highlights how the source conveys significant concern while the reply seeks clarification but lacks emotional depth.
In the ``Deny'' stance, the source text poses a question regarding the truth of Earth's flatness and the existence of proof for this belief. The reply text emphatically denies the validity of the claim, asserting that there is no proof and labeling those who believe it as trolls.
The ``Comment'' stance example regarding nostril breathing features a source text that invites others to share their experiences about their ability to breathe through both nostrils. This source text conveys a positive emotional context, reflecting curiosity and openness. In contrast, the reply provides personal insights but does not express any particular emotional response, offering only information about individual experiences, as mentioned in Table \ref{table:example}.

\subsection{Existing Approaches and Challenges}
Detecting stance in user-generated content is crucial for combating misinformation. While numerous approaches have been proposed in the literature, effectively modeling the interactions between source and reply texts remains a significant challenge, particularly in multi-turn conversations.
Early work in stance detection primarily relied on Convolutional Neural Networks-based approaches to capture textual features and sequential dependencies \cite{9178321, 8851136}. Although these methods showed potential in identifying stance within individual texts, they often struggled with contextual understanding, especially in conversational settings where reply texts are contingent on preceding messages and the target in the source text is implicit. 
Subsequent studies explored attention mechanisms, such as Hierarchical Attention Networks (HANs) \cite{Sun2019} and scaled dot product attention \cite{tata} to better capture contextual dependencies in data. However, they often overlooked the bidirectional inter and intra relationship between source and reply texts. The lack of explicit modeling of inter-textual relationships limited their effectiveness in multi-turn discourse. 

Recent works have investigated the joint modeling of sentiment and stance, recognizing that affective features (e.g., emotion, sentiment, tone) provide critical context for stance interpretation \cite{10098101, electronics10111348}. Works such as Sun \textit{et al}. \cite{Sun2018} and Huang \textit{et al}. \cite{app14093916} demonstrated that sentiment-aware models improve stance detection by capturing the emotional undercurrents of argumentative texts. However, these approaches often treat emotion as a supplementary feature rather than a relational signal between source and reply. This limits their ability to model scenarios where emotional alignment between texts serves as a stance indicator.

\subsection{Research Objectives}
Building upon the identified gaps in stance detection, particularly the challenges of attention mechanisms in capturing implicit intent in source and reply text, as well as affective-feature modeling, our work establishes primary objectives as follows:
\begin{itemize}
    \item {To explore a dual cross-attention mechanism that addresses the complexities of misinformation by enhancing the contextual understanding between source and reply text for stance detection in user-generated content.}
    \item {To investigate the influence of emotion features for capturing emotional alignment between source and reply texts using distance metric learning.}
    \item { To assess whether the similarity or divergence between source and reply texts, computed using distance-metric learning by understanding the contextual relationships between source and reply text, enhances the stance detection.}
 \item {To develop a unified framework using label information in the training phase that improves the stance classification.}
\end{itemize}
\subsection{Key Contributions}
We summarize our key contributions as follows:
\begin{enumerate}[label=\arabic*)]
    \item We propose an emotion-aware dual cross-attentive neural network with label fusion for stance detection. By recognizing that affective features can shape frameworks and responses, the proposed method enhances the understanding of emotions in stances. The label fusion technique integrates label information for effective mapping of the features to specific stance labels, leading to more accurate and contextually
    aware classification. 
    \item We devise a dual cross-attention mechanism for the input texts, followed by a hierarchical attention network to capture inter and intra-relationships. This helps in identifying the important parts of both source and reply texts. 
    \item To explore the impact of emotional alignment of source-reply text on stance, we incorporate emotions expressed in both source and reply texts.
    \item We integrate distance-metric learning to improve its performance in classifying stance. We measure the proximity of various features, including emotional alignment, transformer-based features, and label information, within our feature enhancement, feature closeness, emotion synthesis and label fusion techniques.  
    \item Extensive experimental results on three datasets show
that our proposed method outperforms current baselines and state-of-the-art methods.

\end{enumerate}

The rest of this article is structured as follows. Section \ref{rel_wrk} presents related work done in the field of stance detection. Section \ref{sec:p_def} provides a detailed definition of the problem. We discuss our methodology in Section \ref{sec:sa_meth}. Subsequently, we show the experimental evaluations in Section \ref{sec:exp_eval} and the discussion in Section \ref{sec:dis}. Finally, Section \ref{con} concludes by summarizing our work.
\section{Related Work}
\label{rel_wrk}
Misinformation and stance detection are highly significant tasks within the field of Natural Language Processing (NLP). We examine previous research on stance classification,
categorizing these studies into three general categories: CNN LSTM-based methods, Transformer-based methods, and LLM-based
methods.

\subsection{CNN LSTM-based Methods} 
{Convolutional Neural Networks and Recurrent Neural Networks, specifically Long Short-Term Memory Networks, have demonstrated strong performance in various text classification tasks by capturing both local and sequential features of input data.
CNN-based methods have been particularly effective in extracting local features and incorporating claim information. Karande \textit{et al.} \cite{karande2021stance} propose CNN architectures to enhance the extraction of local features, which are crucial for understanding stances. Rashed \textit{et al.} \cite{rashed2021embeddings} explores a multilingual universal sentence encoder based on CNNs, which projects user-generated text into an n-dimensional embedding space. This approach provides richer semantic representations, thereby improving the model's ability to identify stances across different languages and cultural contexts.
LSTM-based approaches, on the other hand, excel at modeling long-range dependencies and contextual relationships in text \cite{fu2022incorporate,santosh2019can,yang2020tweet}. }
{A notable advancement was introduced by Pu \textit{et al.} \cite{electronics14010186}, which combined BiLSTM with transformer-based multi-task learning to integrate emotion features into stance detection. These approaches overlook affective features in conversational targets (source texts), limiting their effectiveness in dialogue-based stance detection. 
Despite their contributions, such approaches face challenges in comprehensive modeling interactions between source and reply texts, particularly in capturing affective and contextual dynamics.}

\subsection{Transformer-based Methods}
{The introduction of transformer architecture by Vaswani \textit{et al.} \cite{vatt} has fundamentally transformed NLP tasks, including stance detection, by enabling approaches to model textual relationships. 
Recent studies have demonstrated the effectiveness of transformer-based models in this domain. Prakash \textit{et al.} \cite{prakash-tayyar-madabushi-2020-incorporating} employ Robustly Optimized Bidirectional Encoder Representations from Transformers (RoBERTa) \cite{liu2019roberta} combined with Term Frequency–Inverse Document Frequency (TF-IDF) technique, while Hanley \textit{et al.} \cite{tata} leverage Decoding-enhanced Bidirectional Encoder Representations from Transformers (DeBERTa) \cite{he2021debertav3}, both achieving remarkable results in the stance classifcation task.
{Similarly, Dar \textit{et al.} \cite{DAR2024112526} apply RoBERTa to generate text representations, combined with stance label embeddings.}
Kawintiranon \textit{et al.} \cite{kawintiranon2021knowledge} enhance stance detection in social media by incorporating background knowledge, though their approach struggled to capture emotional and relational dynamics between source and reply texts.} 
{In parallel, recent progress in attention mechanisms has shown promise across various tasks \cite{SINGH2025127292, RAGHAW2024108821}.}
{For instance, Rehman \textit{et al.} \cite{REHMAN2025103895} propose a multimodal framework using cross attention but restricted feature interactions to a single attention pass. 
Likewise, Liu \textit{et al.} \cite{liu2024moeit} employ asymmetric Multi-Head Attention (MSA) but neglect inter and intra-modal attention, a critical component for textual understanding. While these methods achieve strong results in their respective domains, they fail to fully exploit cross attention along with self-attentive feature interactions, particularly the bidirectional dependencies between source and reply texts.
In a related direction, Li \textit{ et al.}  \cite{LI2024111457} propose a method for sarcasm detection that incorporates attention mechanisms and emotion alignment, integrating emotions alongside other features. However, their approach considers only a limited range of emotions and provides minimal analysis of emotion dynamics between the source and reply texts.
}
{To address these gaps, SPLAENet introduces a dual cross-attention mechanism that jointly processes source and reply texts, capturing both inter and intra-textual relationships for richer feature representations. Additionally, it incorporates an emotion synthesis module to model emotion alignment between the source and reply texts.}

\subsection{LLM-based Methods} 
Recent advancements in Large Language Models (LLMs) have substantially influenced stance detection methodologies within NLP. Initial approaches demonstrated that LLMs could surpass traditional benchmarks, with Zihao \textit{et al.} \cite{he2022infusing}  combining Wikipedia-derived knowledge, BERT features, and Chat Generative Pre-trained Transformer (ChatGPT) insights to achieve significant improvements. Additionally, Hardalov \textit{et al.} \cite{hardalov2022few} introduce a prompt-based framework that employs a Cross-Lingual Language Model based on the RoBERTa architecture for feature extraction. On the other hand, our method extracted features using RoBERTa-Large.
The field has since evolved toward more sophisticated integration strategies.
{Work by Zhang \textit{et al.} \cite{zhang2022would} demonstrates that combining multiple LLMs enhances performance by aggregating diverse knowledge sources and reasoning perspectives. Lan \textit{et al.} \cite{Lan_Gao_Jin_Li_2024}, a collaborative agent system where specialized LLMs (linguistic expert, domain specialist, and social media veteran) provide multi-faceted analysis before final stance determination.
Zhang \textit{et al.} \cite{zhang-etal-2024-llm-driven} propose injecting LLM-extracted target-text relational knowledge into Bidirectional and Auto-Regressive Transformers (BART) while utilizing prototypical contrastive learning for label alignment.
While LLMs provide broad world knowledge, they potentially miss the stance-specific relationships between source and reply texts due to their general-purpose training objectives. In contrast, SPLAENet integrates emotion analysis, the method infers underlying intents and affective features, enabling robust stance detection even when lexical similarity alone is inadequate.
Our approach advances the field by unifying contextual, relational, and emotional insights for a more comprehensive understanding of stance.}
\section{Problem Definition}
\label{sec:p_def}
Let D represent a dataset containing C number of samples and 
D = \{$T_s^i$, $T_r^i$\} signifies texts from the $i$th sample with $T_s^i$ and $T_r^i$ representing the source and reply text, respectively. The objective is to classify the given post into one of four categories: support, query, deny, and comment. Let $Y_i \in $ \{0, 1, 2, 3\} be the label of the $i$th post, where 0 represents support, 1 signifies query, 2 signifies deny, and 3 indicates a comment. 
Our goal is to develop a method $\mathcal{F}(\mathcal T)$  that takes text $T_s$ and $T_r$ as input and outputs a probability distribution for each class, indicating the likelihood of a post belonging to each category based on its stance.
The task can be formalized as predicting the label for the $i$th post, where
\[
Y_i' = \underset{l \in L}{\arg\max} \ P(Y_i = l \mid \mathcal{T})
\] with $Y_i'$ representing the predicted label for the $i$th post. The operation ${\arg\max}_{l \in L}$ denotes finding the class that maximizes the probability, $L$ represents the maximum number of classes, and
\[P(Y_i = l \mid \mathcal T )\]
denotes the conditional probability of the $i$th post belonging to class $l$ given the text $\mathcal{T}$.

\section{Methodology}
\label{sec:sa_meth}

\begin{figure*}[!h]
    \centering
    \captionsetup{justification=justified}   \includegraphics[width=16cm, height=10cm]{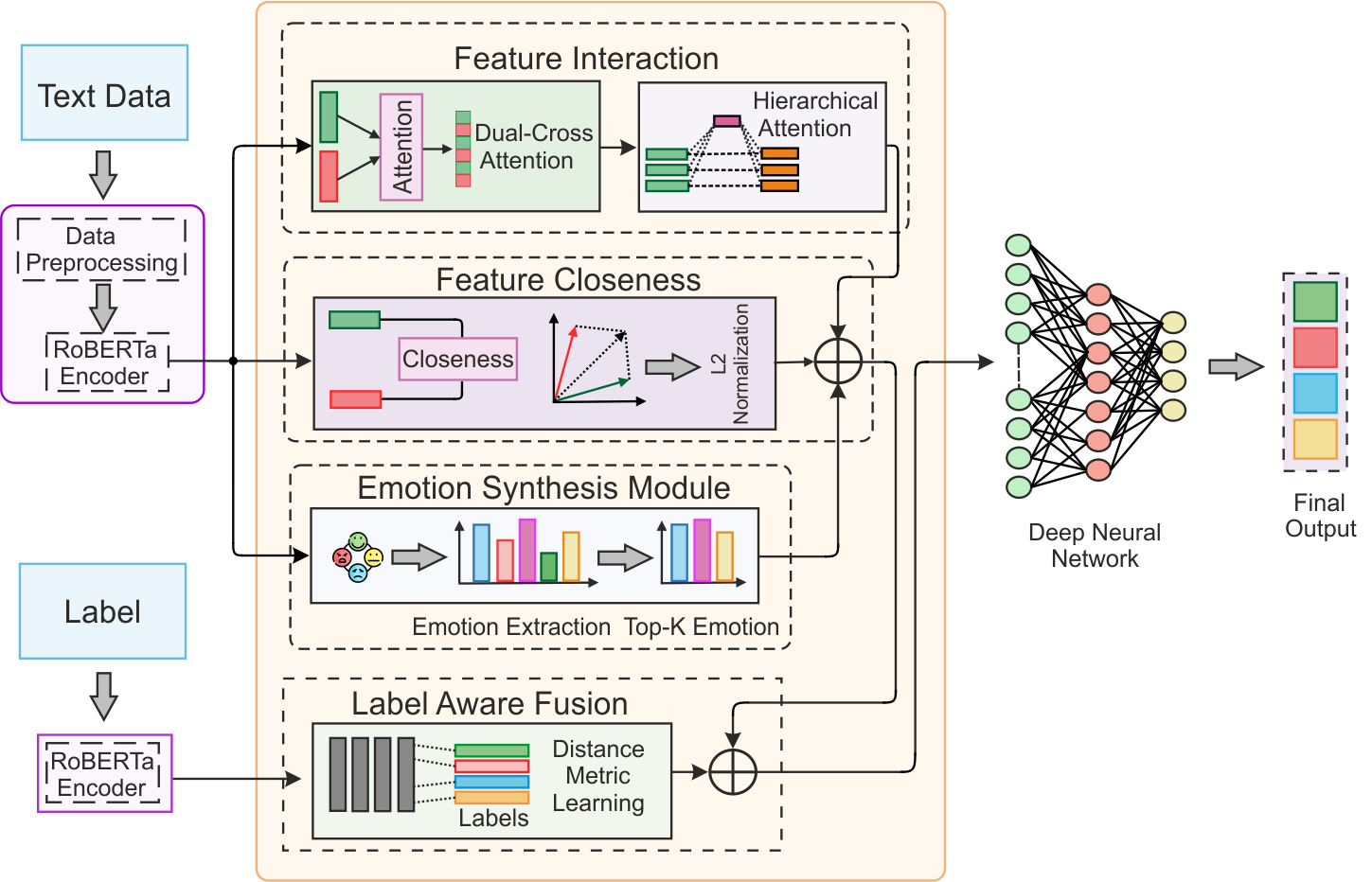}
  
    \caption{ The flow diagram of \textbf{SPLAENet}}
    \label{fig:high_lvl_architecture}
\end{figure*}

This section includes a detailed breakdown of the architecture of the proposed method, as shown in Figure \ref{fig:high_lvl_architecture}. The system takes two user-generated texts as input and aims to identify the underlying stance expressed in the reply text relative to the source text.

\begin{figure*}[!h]
    \centering
    \captionsetup{justification=justified}
    \includegraphics[width=16cm, height=11cm]{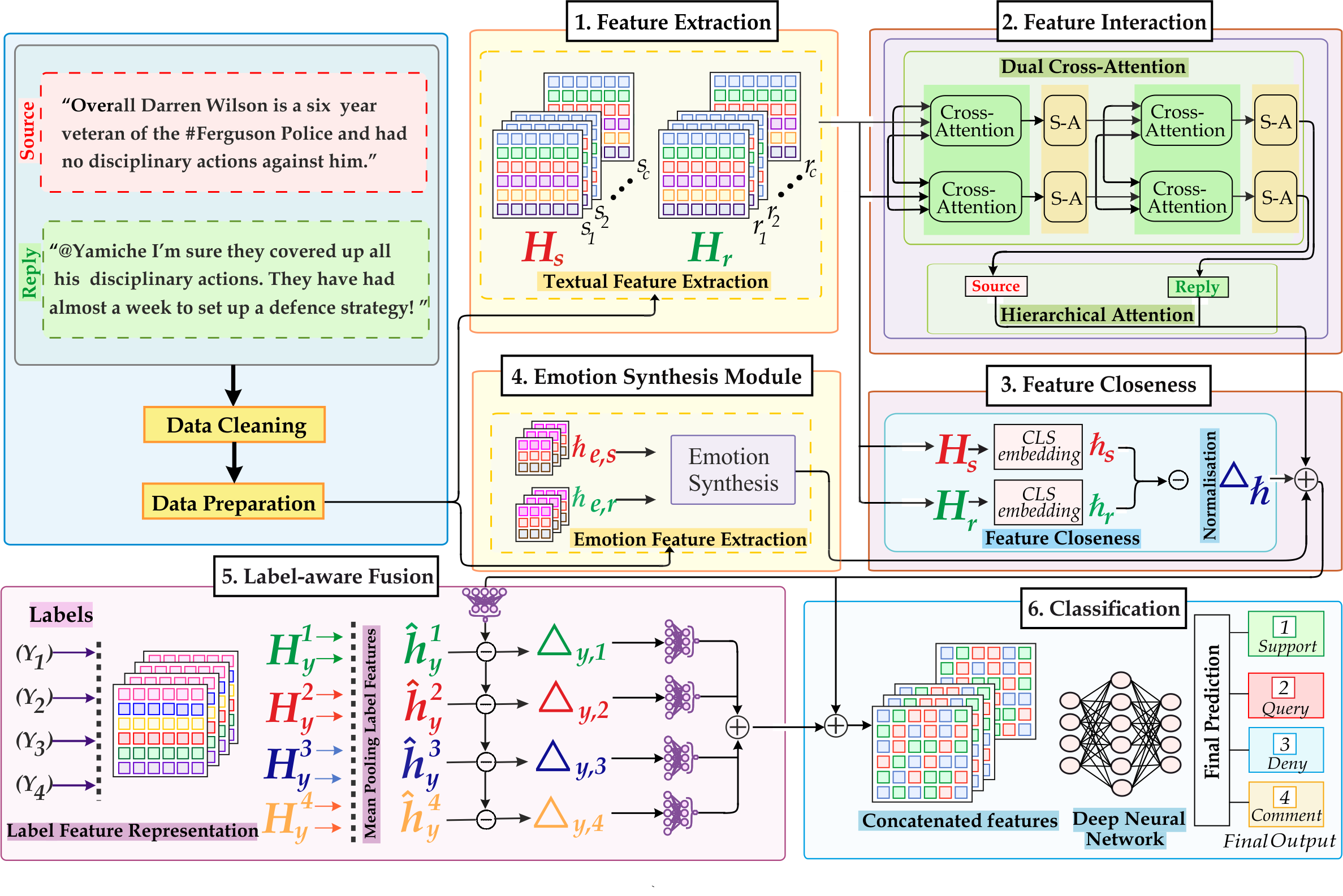}
  
    \caption{ The illustration of \textbf{SPLAENet} - Stance Prediction through a Label-fused dual cross-Attentive Emotion-aware neural Network. The proposed methodology consists of six distinct stages: 
    \textbf{1. Feature Extraction}: In this stage, we extract textual features, denoted as $H_s$, $H_r$, from the respective input text $T_s$ and $T_r$. \textbf{2. Feature Interaction}: Textual features $H_s$ and $H_r$ are processed through our attention module to capture the intricate relationships between $T_s$ and $T_r$. 
    \textbf{3. Feature Closeness}: Proximity between textual features $H_s$ and $H_r$ are evaluated by analyzing their feature representation.
    \textbf{4. Emotion Synthesis Module}: Emotion-specific features ${\hat{\hslash}_{e,s}}$, ${\hat{\hslash}_{e,r} }$ are extracted from $T_s$ and $T_r$ text. The emotion synthesis module facilitates the alignment of emotions between the textual inputs.
    \textbf{5. Label Fusion}: In this phase, the concatenated feature set is integrated with label embedding vectors, enabling the model to learn the proximity of the correct labels relative to their corresponding textual representations. 
    \textbf{6. Classification}: Finally, we classify the resulting feature embedding vectors using dense layers to predict the stance, resulting in a multi-class output aligned with the task requirements.}
    \label{fig:architecture}
\end{figure*}

The proposed method illustrated in Figure \ref{fig:architecture} comprises six major components, namely: 1) feature extraction; 2) feature interaction; 3) feature closeness; 4) emotion synthesis module; 5) label-aware fusion and 6) classification. Initially, the system extracts relevant and contextual features to create informative representations of both source and reply texts.  
 To further enrich the understanding of these texts, emotions are extracted as an additional modality.
We enhance the quality of text representations by leveraging a dual cross-attention mechanism followed by a hierarchical attention network to capture inter and intra-relationships between source and reply texts. For a comprehensive feature representation, the processed emotion features are concatenated with the attended source and reply text features, as well as features capturing the distance between source and reply texts. We employ a label fusion module to gain insights into the proximity of features to all labels. Finally, these features are then fed into a deep neural network to predict the stance.
 The key notations used throughout this paper are summarized in Table \ref{table:notation}.

\begin{table}[!ht]
    \centering
    \captionsetup{justification=raggedright} 
    \caption{Main Notations Used in the Paper}
    \label{table:notation}
    \begin{tabular}{lll}
    \toprule
    \textbf{Symbol}       & \textbf{Dimension}              & \textbf{Description} \\ 
    \midrule
    $d_{\text{model}}$    & 1024                            & Feature size \\
    $T_s$, $T_r$          & $C$                             & Set of source and reply texts \\
    $T_{s}^{\prime}$, $T_{r}^{\prime}$ & $C$                & Tokenized source and reply texts \\
    $U$                   & 50                              & Sequence length \\
    $e_{s}$, $e_{r}$      & $K$                             & Emotions extracted from source and reply texts \\
    $Y$                   & $L$                             & Classification labels \\
    $H_s$, $H_r$          & $\mathbb{R}^{C \times U \times d_{\text{model}}}$ & Hidden layer feature representations of $T_s$ and $T_r$ \\
    $\hslash_s$, $\hslash_r$ & $\mathbb{R}^{C \times d_{\text{model}}}$ & Classification (CLS) token feature representations of $T_s$ and $T_r$ \\
    $\hslash_{e,s}$, $\hslash_{e,r}$ & $\mathbb{R}^{C \times d_{\text{model}}}$ & CLS token feature representations of $e_{s}$ and $e_{r}$ \\
    $\hat{\hslash}_{e,s}$, $\hat{\hslash}_{e,r}$ & $\mathbb{R}^{C \times d_{\text{model}}}$ & Averaged emotion feature representations of top-K $e_{s}$ and $e_{r}$ \\
    $H_{y}$               & $\mathbb{R}^{L \times 3 \times d_{\text{model}}}$ & Hidden layer feature representation of stance label $Y$ \\
    $\hat{H}_{y}$         & $\mathbb{R}^{L \times d_{\text{model}}}$ & Averaged feature representation of stance labels $Y$ \\        
    ${\mathcal{C}}^{\text{Att-m}}_s$, ${\mathcal{C}}^{\text{Att-m}}_r$ & $\mathbb{R}^{C \times U \times d_{\text{model}}}$ & $m$-th cross-attention attended vectors; $m \in \{1,2\}$ \\
    ${\mathcal{S}}^{\text{Att-m}}_s$, ${\mathcal{S}}^{\text{Att-m}}_r$ & $\mathbb{R}^{C \times U \times d_{\text{model}}}$ & $m$-th self-attention attended vectors; $m \in \{1,2\}$ \\
    $Q$, $K$, $V$         & $\mathbb{R}^{C \times U \times d_{\text{model}}}$ & Query, key, and value matrices \\
    $d_k$                 & $\mathbb{R}^{C \times U \times d_{\text{model}}}$ & Dimension of matrix $V_s$ and $V_r$ \\
    $v_s$, $v_r$          & $\mathbb{R}^{C \times d_{\text{model}}}$ & Hierarchical attended context vectors \\
    $\Delta_{E}$          & $\mathbb{R}^{C \times d_{\text{model}}}$ & Difference emotion vector of $\hat{\hslash}_{e,s}$ and $\hat{\hslash}_{e,r}$ \\
    $\Delta_{\hslash}$    & $\mathbb{R}^{C \times d_{\text{model}}}$ & Difference textual vector of $\hat{\hslash}_{s}$ and $\hat{\hslash}_{r}$ \\
    $\tilde{\Delta}_{\hslash}$ & $\mathbb{R}^{C \times d_{\text{model}}}$ & L2 normalization of $\Delta_{\hslash}$ \\
    $f_{\text{cnct}}$     & $\mathbb{R}^{C \times d_{\text{model}}}$ & Concatenated vector with $v_s$, $v_r$, $\Delta_E$, and $\tilde{\Delta}_{\hslash}$ \\
    $f_{\text{fsd}}$      & $\mathbb{R}^{C \times d_{\text{model}}}$ & Concatenated vector of $f_{\text{cnct}}$ and label-specific information \\
    \bottomrule
    \end{tabular}
\end{table}

 \subsection 
{\textit{Feature Extraction}} 
\label{section:social_context_feature}
In this section, we outline the feature extraction module, as illustrated in the architecture depicted in Figure \ref{fig:architecture}. Given that our data consists of textual content, we utilize emotion feature extraction in conjunction with textual features, employing NRC Emotion Lexicon (NRCLex) \cite{mohammad-turney-2010-emotions} for emotions and RoBERTa \cite{liu2019roberta} for textual features. The text features extracted from RoBERTa capture the overall semantics of text, while NRCLex contributes valuable emotional insights.
\subsubsection{\textit{Textual Feature Extraction}}%
Feature extraction entails converting raw text into numerical representations that can be effectively used for further analysis.
To extract textual features from user-generated content, we employ a
transformer-based deep learning model, RoBERTa \cite{liu2019roberta}. 
It leverages several key strategies and techniques to enhance its performance and effectiveness in NLP tasks, including the use of larger training datasets and dynamic masking. We use RoBERTa-Large tokenizer to generate token sequences $T'_s$ and $T'_r$, as shown in Equation (\ref{eq:rob_s}).
\begin{equation}
T'_s = \text{RoBERTa-Large\_Tokenizer}(T_s)
\label{eq:rob_s}
\end{equation}
Here, $T_s$ and $T_r$ represents the source text and reply text, respectively. The preparation of text data for model input is a crucial step.
To maintain uniformity in various text lengths, the maximum token sequence length, U, is limited to 50 for both texts. After conducting experiments with distinct sequence lengths, the limit is set to 50, since the average text length is less than this number.
Shorter sequences are padded with zeros, while longer sequences that surpass the maximum length are truncated. Next, we obtain contextualized embeddings for input texts $T_s$ and $T_r$ using RoBERTa. Each token is processed in the sequence utilizing its encoder, self-attention, and feed-forward network to generate ${d_{\text{model}}}$-dimensional feature vector for each token, as shown in Equation (\ref{eq:H_s}).

\begin{equation} H_{s} = \text{RoBERTa-Large}(T'_s) \label{eq:H_s} \end{equation}
Here, $H_s$ represents the embedding vector which processes the tokenized source text $T'_s$. Similarly, $H_r$ represents the embedding vectors of tokenized reply text $T'_r$.
We extract $H_s$ and $H_r$ from the method's output, which provides contextualized embedding vectors of source and reply texts. By capturing the contextual information and semantic interpretation of the input sequences, these embeddings enable the approach to comprehend and analyze texts.

\subsection{\textit{Feature Interaction}} 
We employ various mechanisms to enable interactions between the textual features of source and reply texts. 
To enhance the textual features, we introduce a dual cross-attention mechanism followed by a hierarchical attention network discussed in Sections \ref{sec:dca} and \ref{sec:wla}, respectively.

\subsubsection{\textit{Dual Cross-Attention}}\label{sec:dca} Attention mechanisms have become a cornerstone of NLP since their introduction by Vaswani \textit{et al.} \cite{vatt}. Attention helps focus on specific parts of a sequence and understand their relationships with other elements. It enables the method to attend to all parts of a sequence simultaneously, facilitating the capture of long-range dependencies crucial for numerous NLP tasks \cite{kumar2020sarcasm,Li2021}.
{Traditional attention-based models, however, primarily focus on intra-textual relationships \cite{sun-etal-2018-stance, zhang-etal-2020-enhancing-cross}. Although advancements such as scaled dot-product attention \cite{tata} and multi-head attention \cite{ conforti-etal-2022-incorporating} enhance the ability to capture dependencies, they often overlook the bidirectional inter and intra-textual relationships between the source and reply texts.
To address the limitations of conventional attention mechanisms, we propose a dual cross-attention mechanism in our framework to enhance the contextual understanding between source and reply texts that focuses on relevant parts of one set of data based on the content of another set of data.  }
\begin{algorithm}[!h]
\caption{Dual Cross-Attention} 
\label{alg:dual_cross_attention}
{\textbf{{Input:} }$H_s \in \mathbb{R}^{C \times U \times {d_{model}}}$ and $H_r \in \mathbb{R}^{C \times U \times{d_{model}}}$} \\
{\textbf{Reconstructed features:} ${MultiHead}_{s} \in \mathbb{R}^{C \times U \times {d_{model}}}, {MultiHead}_{r} \in \mathbb{R}^{C \times U \times {d_{model}}}$}
\begin{algorithmic}[1]
\Function{CrossAttention}{${X}_s, {X}_r, \text{mode}$}
    \State ${Q}_s, {K}_s, {V}_s \Leftarrow {X}_s \cdot {W}_{q,s}, {X}_s \cdot {W}_{k,s}, {X}_s \cdot {W}_{v,s}$
    \State ${Q}_r, {K}_r, {V}_r \Leftarrow {X}_r \cdot {W}_{q,r}, {X}_r \cdot {W}_{k,r}, {X}_r \cdot {W}_{v,r}$
    \For{$i \Leftarrow 1$ to ${num\_heads}$}
        \State ${Q}_{s}^i, {K}_{s}^i, {V}_{s}^i \Leftarrow {Q}_s[:, i, :], {K}_s[:, i, :], {V}_s[:, i, :]$
        \State ${Q}_{r}^i, {K}_{r}^i, {V}_{r}^i \Leftarrow {Q}_r[:, i, :], {K}_r[:, i, :], {V}_r[:, i, :]$
        \If{$\text{mode} == \text{``key''}$}
            \State ${head}_{s}^{i} \Leftarrow \text{softmax}\left(\frac{{Q}_{s}^i ({K}_{r}^i)^T}{\sqrt{\text{depth}}}\right) \cdot {V}_{s}^i$
            \State ${head}_{r}^{i} \Leftarrow \text{softmax}\left(\frac{{Q}_{r}^i ({K}_{s}^i)^T}{\sqrt{\text{depth}}}\right) \cdot{V}_{r}^i$
        \ElsIf{$\text{mode} == \text{``value''}$}
            \State ${head}_{s}^{i} \Leftarrow \text{softmax}\left(\frac{{Q}_{s}^i ({K}_{s}^i)^T}{\sqrt{\text{depth}}}\right) \cdot {V}_{r}^i$
            \State ${head}_{r}^{i} \Leftarrow \text{softmax}\left(\frac{{Q}_{r}^i ({K}_{r}^i)^T}{\sqrt{\text{depth}}}\right) \cdot {V}_{s}^i$
        \EndIf
    \EndFor
    \State ${MultiHead}_{s} \leftarrow \text([{head}_{s}^1 \oplus {head}_{s}^2\oplus \ldots \oplus{head}_{s}^{num\_heads}])$
    \State ${MultiHead}_{r} \leftarrow \text([{head}_{r}^1\oplus {head}_{r}^2\oplus \ldots\oplus {head}_{r}^{num\_heads}])$
    \State \text{return} ${MultiHead}_{s}, {MultiHead}_{r}$
\EndFunction

\Function{SelfAttention}{${X}$}
    \State ${Q}, {K}, {V} \Leftarrow {X} \cdot {W}_{q}, {X} \cdot {W}_{k}, {X} \cdot {W}_{v}$
    \For{$i \Leftarrow 1$ to ${num\_heads}$}
        \State ${Q}^i \Leftarrow {Q}[:, i, :]$
        \State ${K}^i \Leftarrow {K}[:, i, :]$
        \State ${V}^i \Leftarrow {V}[:, i, :]$
        \State ${head}_{i} \Leftarrow \text{softmax}\left(\frac{{Q}^i ({K}^i)^T}{\sqrt{\text{depth}}}\right) \cdot{V}^i$
    \EndFor
    \State ${MultiHead} \leftarrow \text([{head}^1 \oplus {head}^2\oplus \ldots \oplus{head}^{num\_heads}])$
    \State \text{return} ${MultiHead}$
\EndFunction

\State ${\mathcal{C}}_s^{Att-1}, {\mathcal{C}}_r^{Att-1} \Leftarrow \text{C{\small {ROSS}}A{\small TTENTION}}({H}_s, {H}_r, \text{``key''})$
\State ${\mathcal{S}}_s^{Att-1} \Leftarrow \text{S{\small ELF}A{\small TTENTION}}({\mathcal{C}}_s^{Att-1})$
\State ${\mathcal{S}}_r^{Att-1} \Leftarrow \text{S{\small ELF}A{\small TTENTION}}({\mathcal{C}}_r^{Att-1})$
\State ${\mathcal{C}}_s^{Att-2}, {\mathcal{C}}_r^{Att-2} \Leftarrow \text{C{\small ROSS}A{\small TTENTION}}({\mathcal{S}}_s^{Att-1}, {\mathcal{S}}_r^{Att-1}, \text{``value''})$
\State ${\mathcal{S}}_r^{Att-2} \Leftarrow \text{S{\small ELF}A{\small TTENTION}}({\mathcal{C}}_s^{Att-2})$
\State ${\mathcal{S}}_s^{Att-2}  \Leftarrow \text{S{\small ELF}A{\small TTENTION}}({\mathcal{C}}_r^{Att-2})$
\end{algorithmic}
\end{algorithm}

The design procedure of attention module is summarized in Algorithm \ref{alg:dual_cross_attention}. The algorithm incorporates two distinct cross-attention layers, each interspersed with self-attention layers. 
{Line 1 presents the Cross-Attention function, which performs attention between two input sequences ${X}_s$ and ${X}_r$, both sharing the same dimensional structure. 
The core of the algorithm revolves around two key functions, namely  Cross-Attention and Self-Attention.
The Cross-Attention function enables bidirectional interaction between source and reference texts through two distinct modes. In the ``key" mode, each input uses its own queries and values but attends to the other's keys to compute attention weights, creating an alignment based on their respective focuses.
By focusing on the key aspects of each text, the model effectively identifies how the source and reply text relate semantically.
In the ``value" mode, the inputs retain their own queries and keys to compute attention weights, but incorporate each other's values.  This method allows the model to uncover direct connections by identifying similarities from source text to reply text and vice versa. 
These exchanges enhance the contextual understanding by enabling the model to consider how each text influences and is influenced by the other.}
We compute $K_s$, $Q_s$, $V_s$, $K_r$, $Q_r$ and $V_r$ matrices for source and reply texts using learned weight matrices ${W}_{q,s}$, ${W}_{k,s}$, ${W}_{v,s}$, ${W}_{q,r}$, ${W}_{k,r}$, and ${W}_{v,r}$, as given in Lines 2 and 3. Line 4 iterates over the number of attention heads. In Lines 5 and 6, for each $head$, extracts the $i$-th head's query, key, and value matrices for the source and reply sequences. Line 7 determines the cross-attention stage-one depending on the variable mode to be ``key''. 
The cross-attention stage one identifies the agreement between the source and reply text and is mathematically formalized in Equation (\ref{eq:ca1s}) and (\ref{eq:ca1r}).

\begin{subequations}
\begin{equation}
{\mathcal{C}}_s^{Att-1} = \frac{\text{softmax}\left (Q_s({H}_s) ) {K_r({H}_r) }^T \right)}{\sqrt{d_k}} V_s({H}_s)
\label{eq:ca1s}
\end{equation}

\begin{equation}
{\mathcal{C}}_r^{Att-1} = \frac{\text{softmax}\left (Q_r({H}_r) ) {K_s({H}_s) }^T \right)}{\sqrt{d_k}} V_r({H}_r) 
\label{eq:ca1r}
\end{equation}
\end{subequations}
Here, inputs to cross-attention stage-one are contextualized embedding vectors of source and reply texts represented as $H_s$ and $H_r$.
In the cross-attention stage one, the key matrix $K_s$ and $K_r$ are swapped. $Q_s$ attends to its value $V_s$ but utilizes $K_r$ of reply, while $Q_r$ attends to its value $V_r$ but employs $K_s$ of source.
We compute cross-attention heads by performing the scaled dot-product attention between $Q_S$ and $K_r$ for the source text, and between $Q_r$ and $K_s$ for the reply text in lines 8 and 9.
Line 10 determines cross-attention stage two when the mode is ``value''.
At a high level, it establishes an initial alignment between the source and reply text, identifying similarities and direct connections such as common topics both texts are referring to. The cross-attention stage two serves as the operation to discover relationships from two different contexts between the source and reply text and is mathematically formalized in Equation (\ref{eq:ca2s}) and (\ref{eq:ca2r}).
\begin{subequations}
\begin{equation}
{\mathcal{C}}_s^{Att-2} = \frac{\text{softmax}\left (Q_s({\mathcal{S}}_s^{Att-1}) ) {K_s(({\mathcal{S}}_s^{Att-1})) }^T \right)}{\sqrt{d_k}} V_r({\mathcal{S}}_r^{Att-1})
\label{eq:ca2s}
\end{equation}

\begin{equation}
{\mathcal{C}}_r^{Att-2} = \frac{\text{softmax}\left (Q_r({\mathcal{S}}_s^{Att-1}) ) {K_r(({\mathcal{S}}_s^{Att-1})) }^T \right)}{\sqrt{d_k}} V_s({\mathcal{S}}_r^{Att-1})
\label{eq:ca2r}
\end{equation}
\end{subequations}
Here, inputs to cross-attention stage-two are self attentive stage one vectors for source and reply texts represented as ${\mathcal{S}}_s^{Att-1}$ and ${\mathcal{S}}_r^{Att-1}$. In cross-attention stage two, the value matrix for source and reply text represented as $V_s$ and $V_r$ is exchanged.
Lines 11 and 12 shows that $Q_s$ attends to its key $K_s$ but utilizes $V_r$ of reply, 
 while $Q_r$ attends to its own key $K_r$ but employs $V_s$ of source.
Lines 15 and 16 show concatenation of $head$ denoted as $\oplus$ to form multi-head attention vectors for source and reply feature vectors.
Line 19 presents self-attention to refine features. Self-attention is applied to outputs of both stages of cross-attention to source and reply texts. The purpose of self-attention is to refine the contextual understanding within reply attended-source text and source attended-reply text by capturing internal dependencies and relationships is formulated in Equations (\ref{eq:sas}) and (\ref{eq:sar}). 
\begin{subequations}
\begin{equation}
{\mathcal{S}}_s^{Att-m} = \frac{\text{softmax}\left (Q_s({\mathcal{C}}_s^{Att-m}) ) {K_s({\mathcal{C}}_s^{Att-m})) }^T \right)}{\sqrt{d_k}} V_s({\mathcal{C}}_s^{Att-m})
\label{eq:sas}
\end{equation}

\begin{equation}
{\mathcal{S}}_r^{Att-m} = \frac{\text{softmax}\left (Q_r({\mathcal{C}}_r^{Att-m}) ) {K_r({\mathcal{C}}_r^{Att-m})) }^T \right)}{\sqrt{d_k}} V_r({\mathcal{C}}_r^{Att-m})
\label{eq:sar}
\end{equation}
\end{subequations}
Here, m $\in \{1, 2\}$ representing the stages of self and cross attention and ${\mathcal{C}}_s^{Att-m}$ and ${\mathcal{C}}_r^{Att-m}$ represents reply attended source text and source attended reply text, respectively.
Incorporating cross-attention as described  improves the method's capacity to comprehend the text by leveraging interactions between different sequences. The layered combination of cross-attention interspersed with self-attention further refines the embedding, ensuring that the method can capture both inter-sequence relationships and intra-sequence details effectively. 

\subsubsection{\textit{Hierarchical Attention}}\label{sec:wla}
Hierarchical attention allows the model to focus on specific segments within each input sequence, offering insights into which parts of the input are most important. After applying inter and intra-attention to capture the context between feature vectors $H_s$ and $H_r$. These vectors are then fed into a multilayer perceptron (MLP) to introduce non-linearity and generate hidden vectors represented $a_{s(i)}$ and $a_{r(i)}$, as described in Equations (\ref{eq:as}) and (\ref{eq:ar}).

\begin{subequations}
\begin{equation}
a_{{s(i)}} = \tanh(\mathcal{S}_{s(i)}^{Att-2} \cdot W_1 + b_1)
\label{eq:as}
\end{equation}

\begin{equation}
a_{{r(i)}} = \tanh(\mathcal{S}_{r(i)}^{Att-2} \cdot W_2 + b_2)
\label{eq:ar}
\end{equation}
\end{subequations}
Here, $\mathcal{S}_{s(i)}^{Att-2}$ and $\mathcal{S}_{r(i)}^{Att-2}$ represents output vectors of dual cross-attention module which captures inter and intra-attention of source and reply text, and $a_{s(i)}$ and $a_{r(i)}$ are hidden representation of $\mathcal{S}_{s(i)}^{Att-2}$and $\mathcal{S}_{r(i)}^{Att-2}$.
Next, we compute the importance of words represented as $w_{s(i)}$ and $w_{r(i)}$, as shown in Equations (\ref{eq:w_s}) and (\ref{eq:w_r}).

\begin{subequations}
\begin{equation}
w_{s(i)} = \text{softmax}(a_{s(i)}^\top \cdot c_{s(i)})
\label{eq:w_s}
\end{equation}

\begin{equation}
w_{r(i)} = \text{softmax}(a_{r(i)}^\top \cdot c_{r(i)})
\label{eq:w_r}
\end{equation}
\end{subequations}
Here, $a_{s(i)}$ and $a_{r(i)}$ represents hidden vectors and $c_{s(i)}$ and $c_{r(i)}$ are  context vectors. 
The similarity is computed between $a_{s(i)}$ and $c_{s(i)}$ as well as between $a_{r(i)}$ and $c_{r(i)}$
The softmax function is then applied to these similarities. 
These weights are subsequently used to calculate textual context vectors $v_{s(i)}$ and $v_{r(i)}$, as demonstrated in Equations (\ref{eq:v_s}) and (\ref{eq:v_r}).

\begin{subequations}
\begin{equation}
v_{s(i)} = \sum_{i=1}^{U} (w_{{s}(i)} \cdot \mathcal{S}_{s(i)}^{Att-2})
\label{eq:v_s}
\end{equation}

\begin{equation}
v_{r(i)} = \sum_{i=1}^{U} (w_{{r}(i)} \cdot  \mathcal{S}_{r(i)}^{Att-2})
\label{eq:v_r}
\end{equation}
\end{subequations}
Here, $v_{s(i)}$ and $v_{r(i)}$ are context vectors for source and reply texts that capture the essential parts of the input that the model focuses on.
They are computed as the weighted sum of normalized weights $w_{s(i)}$ and $w_{r(i)}$ with the corresponding intra-attention output vectors $\mathcal{S}_{s(i)}^{Att-2}$ and $\mathcal{S}_{r(i)}^{Att-2}$. 

\subsection{\textit{Feature Closeness}}\label{sec:fc}
In the context of the proposed method, feature closeness refers to a technique that quantifies the similarity or dissimilarity between source and reply texts \cite{ramesh2022automated}. It is crucial to evaluate the relationship between the source and reply accurately by analyzing their feature representations. To capture this closeness, we compute the element-wise absolute difference between feature representations of the source and reply text.
To ensure that features are on a uniform scale, L2 normalization is applied to the resulting vector. This Distance-Metric Learning (DML) approach facilitates the assessment of the contextual proximity of two texts based on underlying context.
For both source and reply texts, we extract the [CLS] token. The element-wise absolute difference between [CLS] tokens is computed to capture the proximity between the source and reply textual features, generating a difference feature vector denoted as ${{\Delta}_{\hslash}}$, as shown in Equation  (\ref{eq:diff_nor_f}a). This difference feature vector quantifies how closely the two textual representations are related. To ensure comparability and stabilize the training process, the difference feature vector is further normalized using L2 normalization, as represented by ${{\tilde \Delta}_{\hslash}}$ in Equation (\ref{eq:diff_nor_f}b).
\begin{subequations} \label{eq:diff_nor_f} \begin{equation} {{\Delta}_{\hslash}} = \left|{\hslash_s} - {\hslash_r}\right| \label{eq:diff
} \end{equation} 
\begin{equation}
{{\tilde \Delta}_{\hslash}} = \frac{{{\Delta}_{\hslash}}}{\|{{\Delta}_{\hslash}}\|_2}
\end{equation}

\end{subequations}
Here, ${\hslash_s}$ and ${\hslash_r}$ represent [CLS] token features of the source and reply texts, respectively. The L2 norm $\|{{\Delta}_{\hslash}}\|_2$ refers to the Euclidean norm, which ensures that the magnitude of the difference vector is scaled appropriately, thus maintaining a consistent range of values during training.
\subsection{\textit{Emotion Synthesis Module}}\label{sec:emo}
\begin{figure*}[!h]
    \centering
    \captionsetup{justification=justified}
    \includegraphics[width=16cm, height=4.7cm]{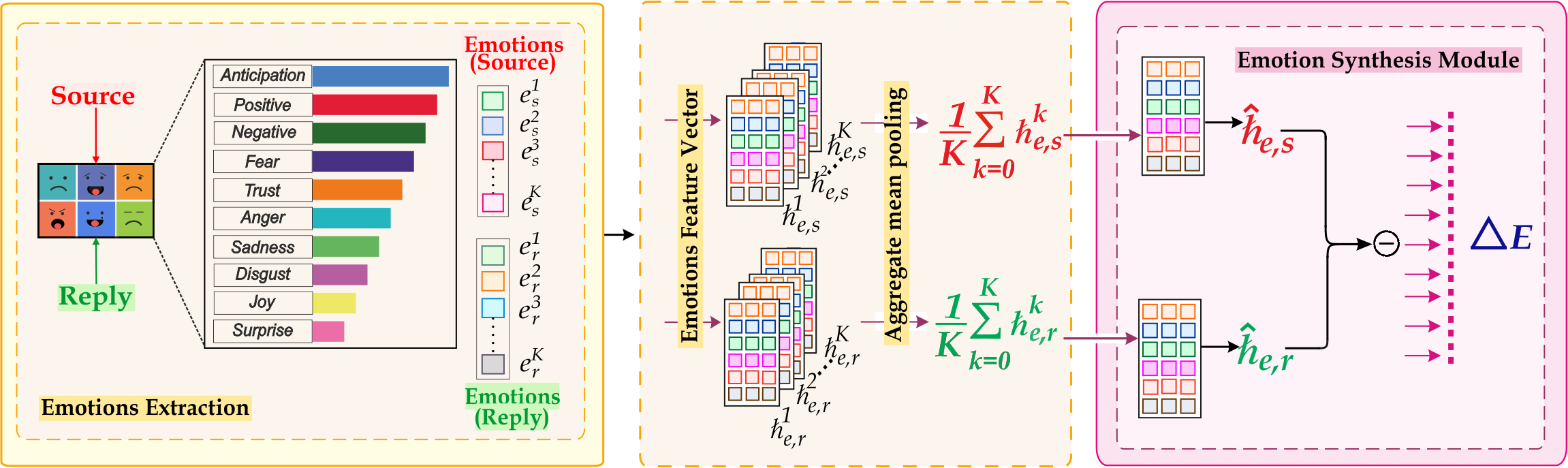}
  
    \caption{Emotion Synthesis Module to capture Emotion Alignment between Source and Reply Text}
    \label{fig:emotion_architecture}
\end{figure*}
Emotions expressed in text help in various text-based downstream tasks such as depression detection \cite{covid-emo}, sentiment analysis \cite{mahaemosen}, and sarcasm detection \cite{babanejad-etal-2020-affective}. The intensity and type of emotions reveal the strength and complexity of the stance. We employ NRCLex for emotion extraction from text.
Textual content from social media contains various emotions expressed by users. These emotions can serve as a pertinent feature in stance detection. They assist in identifying the intent expressed in text. By incorporating emotions into textual analysis, we can more accurately decipher the complexities of stance. 
The proposed approach considers the alignment and pattern of emotions \cite{wheel} between source and reply texts to identify stance. 

Given a text, NRCLex returns a list of 10 emotions; \textit{<fear, anger, anticipation, trust, surprise, positive, negative, sadness, disgust, joy>}, each assigned a score. Our observations reveal a consistent pattern of emotions across source and reply texts for different stances. For instance,
 in the case of the ``Support'' stance, the source predominantly shows emotions such as \textit{trust} and \textit{positive}, while the reply reflects mostly a \textit{positive} emotion. Similarly, when the stance is ``Deny,'' most of the source text exhibit emotions such as \textit{fear, negative, anger,} and \textit{trust}, while replies display emotions such as \textit{fear, disgust, anticipation, negative, sadness,} and \textit{surprise}. For ``Comment'', stance, the source mostly exhibits \textit{positive, fear,} and \textit{trust} emotions, while the reply does not exhibit any emotional engagement in most cases.
 Finally, in ``Query'' stance, the source text displays a high level of \textit{negative}, \textit{fear,} and \textit{trust} emotions, while the replies again demonstrate a lack of emotional content.
Leveraging these insights, we incorporate emotion analysis to enhance stance detection illustrated in Figure \ref{fig:emotion_architecture}.

\begin{figure*}[!h]
    \centering
    \includegraphics[width=\textwidth]{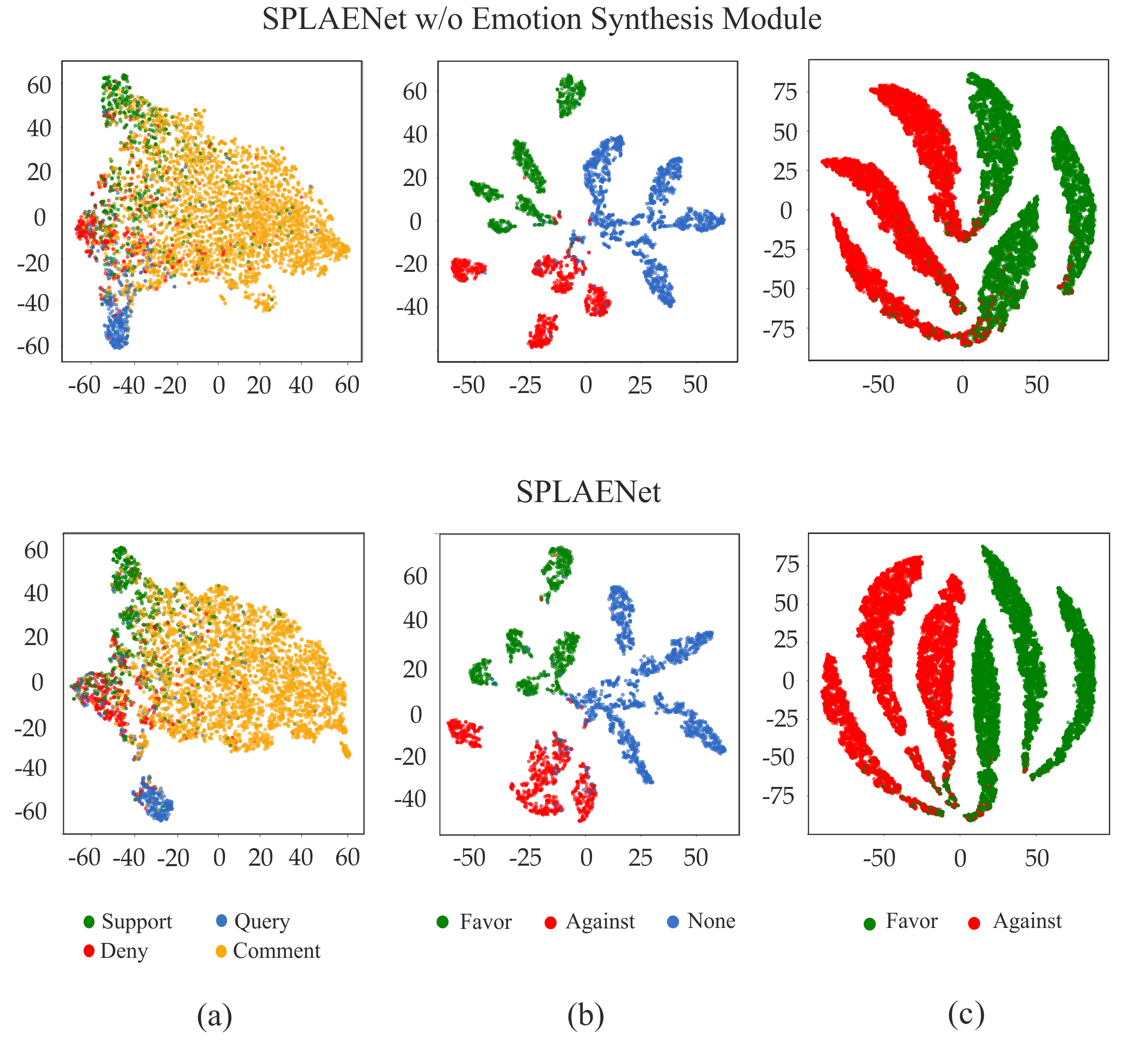}      
        \caption{{The t-SNE Visualization of Intermediate Representations from SPLAENet of Datasets with and without Emotion Synthesis (a) RumourEval (b) SemEval and (c) P-Stance}. Color Map: \textcolor{green}{Support}, \textcolor{bluee}{Query}, \textcolor{red}{Deny}, \textcolor{yellow-or}{Comment}, \textcolor{green}{Favor}, \textcolor{red}{Against}, \textcolor{bluee}{None}}
    \label{fig:comparison}
\end{figure*}

In this process, we extract emotions from source and reply texts  $T_s$ and $T_r$ by creating a dictionary of emotions \cite{mohammad-turney-2010-emotions}.
Emotions are arranged in descending order based on their intensity scores, which aids in selecting the top $K$ emotions represented as $e_s$ in Equation (\ref{eq:nrclex
}). Empirical studies have shown that setting the hyperparameter $K$ to 3 is most effective.

\begin{equation} {e}_{s} = \text{NRCLex}({T_s}) \label{eq:nrclex
} \end{equation}
We then extract [CLS] token feature vectors $\hslash_{e,s}$ and $\hslash_{e,r}$ using RoBERTa encoder to encapsulates the emotional context of the
source and reply text, illustrated in Equation (\ref{eq:rob_emo
}).

\begin{equation} { \hslash_{e,s} }= \text{RoBERTa-Large}(e_s) \label{eq:rob_emo
} \end{equation}
Here, $e_s$ and $e_r$ represent the set of emotions of source and reply texts. These vectors capture the emotional characteristics of source and reply texts. 
The consolidated emotion feature representations of top-$K$ emotions is calculated in Equation (\ref{eq:emo_pool}), represented as ${\hslash_{e,s}}$ and ${\hslash_{e,r}}$,
\begin{equation} 
\hat \hslash_{e,s}= \frac{1}{K}\sum_{k=1}^{K} {\hslash_{e,s}^{k}} \label{eq:emo_pool}
\end{equation}
Here, $K$ denotes the number of top emotion embedding vectors selected, and ${\hslash_{e,s}^{k}}$ represents $k$-th emotion embedding vector of $e_s$. The consolidated emotion feature $\hat{\hslash}_{e,r}$ for reply text is computed by applying the same averaging operation to its top-$K$ emotion embedding vectors.
These consolidated vectors encapsulate the overall emotional representation of both source and reply texts.
To quantify emotional divergence between source and reply texts, we perform element-wise absolute subtraction between aggregated mean pooled vectors and emotion feature vectors represented by $\Delta_{E}$, mathematically expressed in Equation~(\ref{eq
}). 
\begin{equation} \Delta_{E} = \left| {\hat{\hslash}_{e,s} } - {\hat{\hslash}_{e,r} } \right| \label{eq
} \end{equation}
Here, $\hat{\hslash}_{e,s}$ and $\hat{\hslash}_{e,r}$ represent emotion feature vectors of  top-$K$ emotions, based on intensity scores from each text.
For instance, a source text expressing anger towards a specific topic is more likely to be met with a reply conveying joy or amusement if the reply stance is against the source's viewpoint. 
Subsequently, feature representations obtained from feature interaction detailed in section \ref{sec:wla}, feature closeness in section \ref{sec:fc} and emotion features in section \ref{sec:emo} are concatenated and denoted as $f_{fsd}$, as shown in Equation (\ref{eq:cnct}).
\begin{equation}
f_{\text{cnct}} = [v_s \oplus v_r \oplus \Delta_{E} \oplus {\tilde{\Delta}_{\hslash}} ]
\label{eq:cnct}
\end{equation}
Here, $v_s$ and $v_r$ represent the attention-applied feature vector, ${\tilde{\Delta}_{\hslash}}$ represents the normalized differential feature vector and $\Delta_{E}$ represents the difference emotion vector respectively, $\oplus$ denotes the concatenation operation between two embedding vectors.
This method enriches the representation of textual data by capturing alignment of emotions, enhancing the method's ability to discern stance variations when the intent is not explicitly conveyed in the text. This approach allows the method to generalize better across diverse texts and domains by integrating emotion features into their representations.
{Figure \ref{fig:comparison} illustrates class discrimination achieved by SPLAENet in final layers across three datasets. Figure \ref{fig:comparison}(a) denotes results on RumourEval dataset, Figure \ref{fig:comparison}(b) on SemEval dataset and Figure \ref{fig:comparison}(c) on P-Stance dataset. The visualization demonstrates how linear separation is better in our method among the four, three and two classes, emphasizing the influence of coupled features within our framework.}

\subsection{\textit{Label Fusion}}\label{sec:laware}
Label fusion is an essential technique for enhancing the method's ability to determine the relative stance between the reply text and the source text. 
{Current approaches \cite{ liang2022zero, zhao-etal-2024-zerostance} treat stance classification as purely text-driven, neglecting the modeling of label proximity within the stance classification context. 
Our label-aware fusion technique integrates stance label representations directly into the learning process. This technique evaluates the proximity between each classification label and contextual features \cite{wang-etal-2018-joint-embedding,yang-etal-2019-deep}. Dar \textit{et al.} \cite{DAR2024112526} explore label-aware representations for the classification task. However, their approach fails to measure proximity between contextual features and label information. To address this, we propose a joint embedding of textual features and stance label representations, enabling the model to leverage both contextual and label-related information across texts.
} 
Label fusion begins by extracting label-specific features using RoBERTa to generate meaningful label-oriented embedding vectors.
RoBERTa-large model, generating a ${d_\text{model}}$-dimensional feature vector $H_l$ for each label is demonstrated in Equation (\ref{eq:label_roberta}).
\begin{equation}
H^{l}_{y} = \text{RoBERTa-Large}({Y}_{l})
\label{eq:label_roberta}
\end{equation}
Here, $Y_l$ is $l$-th stance label and $l \in {1, 2 \cdots L}$, which is fed to RoBERTa-large. Then they are aggregated and mean-pooled for dimension reductionalilty and represented as $\hat{h_y^l}$.
The process for fusing labels with the concatenated features is outlined in Algorithm \ref{algo:label-fusion}.
The concatenated textual representation from source and reply texts is presented as $f_{cnct}$ are then compared with these label-specific features $\hat{h_y^l}$. By subtracting label features from concatenated features and computing element-wise absolute differences, the technique calculates proximity metrics for each label. These proximity metrics reveal how closely each label aligns with the reply text, enhancing the method's ability to discern relationships between source and reply texts.
\begin{algorithm}[!h]
\caption{Label Fusion}
\label{algo:label-fusion}
\textbf{Input:} $f_{\text{cnct}}$ and $\hat{{h}}_{y}^l$ \\
\textbf{Output:} $f_{\text{fsd}}$
\begin{algorithmic}[1]
\Function{Label Fusion}{}
    \State $f'_z \gets [\ ]$
    \State $\tilde{z} \gets w \cdot f_{\text{cnct}} + b$   
    \For{$l = 1$ to $L$}
        \State $\Delta_{y,l}\gets |\tilde{z} - \hat{h}_{y}^{l}|$
        \State $\tilde{z}_l \gets \sum_{l} w_1 \cdot \Delta_{y,l} + b_1 \in \mathbb{R}^{512}$
        \State $\tilde{z}_l' \gets \sum_{l} w_2 \cdot \tilde{z}_l + b_2 \in \mathbb{R}^{256}$
        \State ${f}_z' \gets {f}_z' \oplus \tilde{z}_l'$
    \EndFor
    \State $f_{\text{fsd}} \gets f_{\text{cnct}} \oplus {f}_z'$
    \State \textbf{return} $f_{\text{fsd}}$
\EndFunction
\end{algorithmic}
\end{algorithm}
Line 2 initializes an empty list $f'_z$, which will store the transformed features for each label $Y_l$.
To integrate label information, we transform \( f_{cnct} \) into a lower-dimensional latent representation $\tilde{z}$ to match the dimension $d_\text{model}$ of label feature ${{H}^{i}_{y}}$, as given in line 3. Line 4 iterates over all L labels. 
Line 5 computes the element-wise absolute difference $\Delta_{y,l}$ between each label embedding vector ${{H}^{l}_{y}}$ and transformed feature $\tilde{z}$ sequentially. Lines 6 and 7 apply linear transformation on $\Delta_{y,l}$ to form $\tilde{z_l}$ and again a linear transformation on $\tilde{z_l}$ to output $\tilde{z_l}'$. Line 8 appends $\tilde{z}_i'$ to $f_z'$, combining transformed features for each label. Line 10 concatenates the original enhanced features $f_{cnct}$ with transformed features $f_z'$ to form the final output feature set $f_{fsd}$.
The label fusion vector \( f_{fsd} \) captures both concatenated features and label-specific information, enhancing the method's ability to discern stance. 
\subsection {\textit{Classification}} 
\label{section:Classification}
The stance classification process is elucidated in this section. The multi-layered processing of extracted features is done through a series of layers, employing both concatenation and convolution operations. The deep neural network processes textual features and emotion features, integrating them with proximity-related and label features through concatenation, then outputs the predicted stance of the reply text concerning the source text. The architecture of the neural network is further outlined in the algorithm below, which details the mathematical flow of the procedure.
The concatenated feature $f_\text{fsd}$ of size 4096 is subsequently passed through multiple fully connected dense layers. This process is demonstrated in Equation (\ref{eq:f0}).
\begin{equation}
 \lambda_i = (W_i \cdot \lambda_{i-1} + b_i)
\label{eq:f0}
\end{equation}
Here, $\lambda_0$ is concatenated feature 
vector $f_{fsd}$, $\lambda_i$ is output feature vector of $i$-th layer, $W_i$ and $b_i$ are weight matrix and bias vector for $i$-th layer and $i \in \{1,2 \ldots \}$. The features are ultimately processed through a dense layer comprising four, three or two labels. A softmax activation function is applied to derive probabilities of labels for given texts. The classification label $Y_l$ is determined by selecting the maximum probability $P$, as shown in Equation (\ref{eq:Pi}).
\begin{equation}
P=max(\text{softmax}(W_n\cdot \lambda_{(n-1)}+b_n))
\label{eq:Pi}
\end{equation}
Here, $\lambda_{(n-1)}$ is the final layer output and $W_n$ and $b_n$ are the weight matrix and bias vector for the final classification layer.
After softmax scaling, the maximum probability is obtained, which can then be indexed to determine the target stance of the given source and reply text. 
\subsection{Workflow of SPLAENet}
{To provide a comprehensive understanding of SPLAENet, we present a detailed pseudo-code in Algorithm \ref{algo:work-flow} that outlines its key stages: textual feature extraction, cross-attention mechanisms, hierarchical attention, feature closeness, emotion synthesis, label fusion, and stance classification. Lines 1 and 2 define the input as dataset $\mathcal{T}$ containing source and reply text pairs and the output as their corresponding stance labels.}
{
Line 3 initiates the iteration over every pair $(t_s^i, t_r^i)$ in the dataset.
Lines 4-6 focus on textual feature extraction where RoBERTa is used to obtain contextualized features $H_s$ and $H_r$.}
{
Dual cross attention mechanism is applied to the embedding $H_s$ and $H_r$ to capture dependencies between source and reply text, resulting in cross attentive representations $\mathcal{S}^{\text{Att-m}}_s$ and $\mathcal{S}^{\text{Att-m}}_r$ in lines 7 and 8.}
{
In lines 9-11, hierarchical attention is performed to produce vectors $v_s$ and $v_r$, which capture the global semantic information of the source and reply texts, respectively.}

{ Feature similarity vector $\tilde{\Delta}_{\hslash}$ is computed between source and reply text in lines 12 and 13.}
{
Lines 14 to 20 focus on emotion feature extraction, where each emotion in the top-K set is encoded using RoBERTa to yield emotion-specific embeddings $\hslash_{e,s}^j$ and $\hslash_{e,r}^j$.
Line 21 fuses all emotion representation into a single vector $\Delta_E$.
Lines 22 and 23 perform feature concatenation, combining hierarchical attention, similarity features and emotion synthesis into one unified vector $f_{cnct}$.}
{
Lines 26-29 begin the label fusion, where stance label names are encoded, and DML computes their alignment with the input representation to produce $f_{fsd}$.}
{Line 31 applies the classifier to the fused features to predict the final stance label. Finally, line 32 returns the predicted stance Y, thereby 
 completing the SPLAENet pipeline.}

\begin{algorithm}[!h]
\caption{{The Workflow of SPLAENet}}
\label{algo:work-flow}
\begin{algorithmic}[1]
    \State \textbf{Input:} Text Dataset $\mathcal{T} = \{(T_s^i, T_r^i)\}_{i=1}^{C}$
    \State \textbf{Output:} Stance Labels $\{Y_i\}_{i=1}^{C}$
    
    \For {each $(t_{s}^{i}, t_{r}^{i})$ in dataset $\mathcal{T}$}
        \State \textbf{Textual Feature Extraction:}
        \State $H_s \gets \text{RoBERTa}(T_s^i)$ \Comment{Source text features}
        \State $H_r \gets \text{RoBERTa}(T_r^i)$ \Comment{Reply text features}
        \State \textbf{Cross Attention:}
        \State ${\mathcal{S}}^{\text{Att-m}}_s, {\mathcal{S}}^{\text{Att-m}}_r \gets \text{DualCrossAttention}(H_s, H_r)$
        
        \State \textbf{Hierarchical Attention:}
        \State $v_s$ $\gets$ \text{HierarchicalAttention}$({\mathcal{S}}^{\text{Att-m}}_s)$
        \State $v_r \gets \text{HierarchicalAttention}({\mathcal{S}}^{\text{Att-m}}_r)$
        
        \State \textbf{Feature Closeness:}
        \State $\tilde{\Delta}_{\hslash} \gets \text{FeatureSimilarity:}(H_s, H_r)$
        \State \textbf{Emotion Synthesis:}
        \State $\{e_s^j\}_{j=1}^K, \{e_r^j\}_{j=1}^K \gets \text{NRCLex}(T_s^i, T_r^i)$ \Comment{Extract top K emotions}
        
        \State \textbf{Emotion Feature Extraction:}
        \For {each emotion pair $(e_s^j, e_r^j)$}
            \State $\hslash_{e,s}^j \gets \text{RoBERTa}(e_s^j)$
            \State $\hslash_{e,r}^j \gets \text{RoBERTa}(e_r^j)$
            
        \EndFor
        \State $\Delta_E \gets \text{Emotion Synthesis }(\hslash_{e,s}^j, \hslash_{e,r}^j)$
        \Comment{Fuse all emotion features}

        \State \textbf{Feature Concatenation:}
        \State $f_{\text{cnct}} \gets [v_s \oplus v_r \oplus \tilde{\Delta}_{\hslash}  \oplus \Delta_E]$
    \EndFor
    
    \State \textbf{Label Fusion:}
    \For {each stance label $L \in \{\text{Support, Query, Deny, Comment}\}$}
        \State $\hat{H}_{y}^L \gets \text{RoBERTa}(L)$ \Comment{Label feature extraction}
        \State $f_{\text{fsd}} \gets \text{DML}(f_{\text{cnct}}, \hat{H}_{y}^L)$ \Comment{Distance Metric Learning}
    \EndFor
    
    \State \textbf{Classification:}
    \State $Y \gets \text{Classifier}(f_{\text{fsd}})$
    
    \State \textbf{Return} $Y$
\end{algorithmic}
\end{algorithm}

\section{Experimental Evaluations}
\label{sec:exp_eval}
This section illustrate the datasets used for the experimental evaluation of our proposed method, SPLAENet, for stance classification. In the following, we present details of the experimental design, including the comparison methods and evaluation metrics used. Finally, the experimental results demonstrate the effectiveness of the proposed technique and its efficacy in stance classification.
\subsection {\textit{Experimental Setup}}
This section begins by outlining hyperparameters used for training the proposed method, followed by a detailed description of the datasets employed in our experimental evaluations. Next, we present an overview of the various methods utilized for comparative analysis. Subsequently, we discuss the evaluation metrics used to assess the performance of different approaches. Finally, we conduct an ablation analysis to examine the impact of different components of the method and conclude with a qualitative analysis.
\subsubsection{\textit{Model Hyperparameters}}\label{label:mod_hyper}
This section describes a systematic evaluation of hyperparameters for the proposed method, concentrating on factors such as number of epochs, batch size, learning rate, and optimizer, to uncover optimal settings that enhance segmentation performance. Training and validation losses, alongside accuracy metrics, are tracked to establish the best number of epochs for model training. 
\begin{table}[ht]
\centering
\caption{An overview of the hyperparameter used for training SPLAENet}
\label{tab:hyperparameters}
{
\begin{tabular}{ll}
\toprule
\textbf{Hyperparameter}       & \textbf{Value} \\ 
\midrule 
Optimizer                     & AdamW                      \\ 
Learning Rate         & $2e-6$                     \\ 
Dropout Rate                  & 0.2                        \\ 
Batch Size                    & 8                           \\ 
Number of Epochs              & 10                          \\ 
Callback            & EarlyStopping    \\ 
\bottomrule
\end{tabular}
}
\end{table}
The recommended hyperparameter configuration is detailed in Table \ref{tab:hyperparameters}.
The pre-trained RoBERTa model is available via Hugging Face repository. We thoroughly investigate the hyperparameters of our proposed model, including learning rate, weight decay, epochs, dropout rate, batch size, and optimizer, aiming to pinpoint the most effective configurations for enhanced training and overall model performance.
For the learning rate and weight decay, we explore values ranging from 
$2e-3$ to $ 2e-7$, finding $2e-6$ to be the most effective. The number of epochs is varied between 5 and 20, with 10 epochs deemed optimal for achieving convergence and improving performance. Our analysis of dropout rates from 0.1 to 0.5 suggests that a rate of 0.2 is optimal, offering effective regularization while maintaining the model's learning capacity. Ultimately, after assessing different optimizer options, such as Adaptive Moment Estimation (Adam), Adaptive Moment Estimation with  Weight Decay (AdamW), and Root Mean Square Propagation (RMSProp), we conclude that AdamW delivers superior performance for our learning process. Thus, the recommended hyperparameter configuration consists of a learning rate of 
$2e-6$, batch size of 8, 10 epochs, a dropout rate of 0.2, and utilization of the AdamW optimizer, balancing model performance effectively.
\subsubsection{\textit{Summarization of Datasets}}\label{label:dataset_summary}

We perform stance detection on three different datasets, RumourEval \cite{gorrell-etal-2019-semeval}, Semeval \cite{mohammad2016semeval}, and P-Stance \cite{p-stance}. All are publicly available English datasets from RumourEval (Task 7a of SemEval 2019), SemEval (Task 6A) and P-Stance. While all tasks involve stance detection, RumourEval focuses more on rumors and their specific characteristics within social media discussions, whereas SemEval addresses stance detection in various social contexts beyond just rumors, and p-stance includes stance towards important events like president elections.
To ensure robust evaluation, we intentionally selected these datasets for their diverse class distribution, where the RumourEval dataset exhibits highly unbalanced, the SemEval shows moderately unbalanced, and the P-Stance maintains relatively balanced label proportions. This variation in label proportions enables a comprehensive assessment of our method's performance across different imbalance scenarios.
Table \ref{table:desc_combined} presents a detailed overview of the RumourEval, SemEval and P-Stance datasets.
\begin{table}[!ht]
\centering
\caption{Dataset Description of RumourEval and SemEval}
\label{table:desc_combined}
\resizebox{\textwidth}{!}{
\begin{tabular}{lccccccccc}
\toprule
\textbf{Label} & \textbf{Percentage (\%)} & \textbf{Count} & 
\textbf{Label} & \textbf{Percentage (\%)} & \textbf{Count}&\textbf{Label} & \textbf{Percentage (\%)} & \textbf{Count} \\
\midrule
\multicolumn{3}{l}{\textbf{RumourEval Dataset}} & \multicolumn{3}{l}{\textbf{SemEval Dataset}}&\multicolumn{3}{l}{\textbf{P-Stance Dataset}} \\
Comment & 75.11 & 6,165 & Favor & 50.68 & 2,110& Favor & 48.36 & 10,431 \\
Support & 11.23 & 819 & Against & 25.39 & 1,057  & Against & 51.64 & 11,143\\
Deny & 7.04 & 567 & None & 23.93 & 996& - & - & - \\
Query & 6.62 & 553 & - & - & -& - & - & - \\
\midrule
\textbf{Total} & \textbf{100} & \textbf{8,083} & \textbf{Total} & \textbf{100} & \textbf{4,163}&\textbf{Total} & \textbf{100} & \textbf{21,574} \\
\bottomrule
\end{tabular}}
\end{table}

\begin{enumerate}

\item{
        \textit{RumourEval}: The RumourEval-2019 dataset (Task 7a of SemEval-2019) \cite{gorrell-etal-2019-semeval} is a critical benchmark for stance detection in the context of online misinformation and rumours, particularly on Twitter. It consists of a total of 8,529 posts, collected from both Twitter and Reddit. Of these, 6,634 are tweets from Twitter, while 1,895 are posts from Reddit. The dataset includes two main types of text: source posts, which are original messages spreading a rumour, and reply posts, which engage with these rumours by expressing a stance. Reply posts offer varied perspectives on the source rumours, classified into four stance categories: Support, Deny, Query, or Comment. Table \ref{table:data_preprocessed} provides a detailed overview of the dataset before and after preprocessing. After preprocessing, the dataset was refined to 8,083 total posts. During data cleaning and preprocessing, certain reply posts are removed as they became null following the elimination of special characters, emojis, and other non-textual elements. Additionally, duplicate source-reply pairs are identified and excluded.
        This dataset is especially valuable for our research because it captures the challenges of noisy, short-text social media data, where stance is often implicit and context-dependent.        
        {However, the dataset has limitations, including its focus on a specific set of rumours, which may restrict the generalizability of models trained on it. Additionally, imbalanced label distributions within the stance categories can introduce bias into model predictions. Despite these challenges, RumourEval-2019 dataset remains a valuable resource for developing and evaluating stance detection models. Its rich annotations and diverse range of posts offer a solid foundation for advancing research in this critical area.}
      
        }
\begin{table}[!ht]
\centering
\caption{Dataset Description of RumourEval Before and After Preprocessing}
\label{table:data_preprocessed}
{
\begin{tabular}{lcccc}
\toprule
\textbf{Split} & 
\multicolumn{2}{c}{\textbf{Before Preprocessing}} & 
\multicolumn{2}{c}{\textbf{After Preprocessing}} \\ 
\cmidrule(lr){2-3} \cmidrule(lr){4-5}
& \textbf{Percentage (\%)} & \textbf{Count} & \textbf{Percentage (\%)} & \textbf{Count} \\ 
\midrule
Train & 75.11 & 5,217 & 50.68 & 4,890 \\ 
Val   & 11.23 & 1,485 & 25.39 & 1,447 \\ 
Test  & 7.04  & 1,827 & 23.93 & 1,746 \\ 
\midrule
\textbf{Total} & \textbf{100} & \textbf{8,529} & \textbf{100} & \textbf{8,083} \\ 
\bottomrule
\end{tabular}}
\end{table}
\item{ 
\textit{SemEval}: SemEval-2016 (Task 6A) dataset \cite{mohammad2016semeval} is a seminal benchmark for target-specific stance detection, where the task is to predict whether a tweet expresses favor, against, or neutral sentiment toward controversial topics (e.g., ``Climate change is real''). It comprises of 4,163 tweets, collected using targeted keyword searches and APIs, focusing on topics such as climate change, feminism, and others related to societal debates. After collecting a large set of tweets, annotators manually label these tweets based on their stance towards the given topic. Its clean annotations and balanced labels make it a standard for comparing against prior work, ensuring our results align with established baselines in the field. Table \ref{table:target-description}(a) presents a target-wise overview
of the SemEval dataset. {The dataset consists of single-turn tweets, lacking conversational context, which limits its ability to capture stance evolution over time. Additionally, it does not account for the diverse ways stance is conveyed, such as implicit or sarcastic expressions. However, this foundational dataset provides a solid base for developing models to explore dynamic stance distribution across diverse topics and evolving time frames.}
\begin{table}[!ht]
\centering
\caption{Target-wise Dataset Description}
\label{table:target-description}
\begin{minipage}{0.45\textwidth}
\centering
\caption*{(a) SemEval Dataset}
\begin{tabular}{lcc}
\toprule
\textbf{Target} & \textbf{Train} & \textbf{Test} \\ 
\midrule
Atheism                     & 513   & 220  \\       
Climate Change is Concern   & 395   & 169  \\    
Feminist Movement           & 664   & 285  \\ 
Hillary Clinton             & 689   & 295  \\    
Legalization of Abortion    & 653   & 280  \\ 
\midrule
\textbf{Total}              & \textbf{2,914} & \textbf{1,249} \\ 
\bottomrule
\end{tabular}
\end{minipage}%
\hspace{1.1cm} 
\begin{minipage}{0.45\textwidth}
\centering
\caption*{(b) P-Stance Dataset}
\begin{tabular}{lccc}
\toprule
\textbf{Target} & \textbf{Train} & \textbf{Val} & \textbf{Test} \\ 
\midrule
Trump   & 6,362   & 795 & 796  \\       
Biden   & 5,806   & 745 & 745  \\    
Sanders & 5,056   & 634 & 635  \\ 
\midrule
\textbf{Total} & \textbf{17,224} & \textbf{2,174} & \textbf{2,176} \\ 
\bottomrule
\end{tabular}
\end{minipage}
\end{table}
}

\item{ 
\textit{P-Stance}: The P-Stance dataset \cite{p-stance} is a large collection primarily designed for stance detection within the political domain, containing 21,574 English-labelled tweets, which is larger and more challenging compared with previous datasets for stance detection. This dataset enables researchers to determine whether a text's author is in favor or against, towards a particular target, such as ``Donald Trump'', ``Joe Biden'', and
``Bernie Sanders'' offers valuable insights into political events, which are collected
during the 2020 United States of America (USA) election. Table \ref{table:target-description}(b) presents a target-wise overview
of the P-Stance dataset. 
{With ~30,000 annotated tweets, it is one of the largest resources for political stance analysis, enabling robust evaluation and scalability. 
The dataset, being more than three times larger than the previous benchmarks \cite{gorrell-etal-2019-semeval,mohammad2016semeval}, poses significant challenges, particularly in handling linguistic complexities. Political discourse often involves implicit bias and sarcasm, making stance detection more difficult. Despite these challenges, the dataset's substantial scale and richness provide valuable opportunities for advancing research in stance detection and political discourse analysis.}
}
\end{enumerate}
\subsubsection{\textit{Data Preprocessing}}  

{Data preprocessing is a crucial step in transforming raw data into a structured and analyzable format, ensuring a strong foundation for robust feature extraction. Our preprocessing pipeline consists of three key phases: } 

\textit{{\textbf{Removing Trailing Hashtags and URLs:}}}  
{
To reduce noise and maintain the discourse structure, we replace all Uniform Resource Locators (URLs) and user mentions in both source and reply texts with special tokens, namely, \$URL\$ and \$MENTION\$. This approach helps to preserve the text's contextual integrity while standardizing its format \cite{fajcik2019but}. By substituting mentions and links, we eliminate potential biases introduced by specific usernames or external web references, ensuring that the model focuses on the core textual content rather than extraneous identifiers. } 

\textit{{\textbf{Data Handling:}}}  
{
Effective handling of missing or redundant data is crucial for ensuring consistent model performance.
During preprocessing, special characters, emojis, and other non-textual elements are removed.
As a result, some reply posts become empty and are subsequently excluded from the dataset. Additionally, replies marked as ``[deleted],'' commonly found in user-generated are also removed as they do not provide meaningful information. Redundant instances, such as duplicated source-reply text pairs, are identified and filtered out to enhance the dataset’s overall quality and uniqueness. Furthermore, entries with missing essential fields, such as the source or reply text, are eliminated, as these components are necessary for effective stance classification.}

{\textbf{\textit{Conversational Structure Processing:}}}
{ Datasets such as RumourEval \cite{gorrell-etal-2019-semeval}, contain threaded conversations where replies consist of multiple sub-replies, forming a hierarchical discourse structure.
To accurately represent the flow of threaded conversations in RumourEval dataset, we use a structured concatenation method. In such datasets, a single reply post \( t_r^i \) may have multiple sub-replies \( \{ t_{r{,j}}^i \}_{j=1}^q \) linked to the same source text \( t_s^i \). To maintain the conversational structure, we sequentially combine each sub-reply with its parent reply. We construct sequences such as \(t_{r}^i t_{r{,1}}^i\), \(t_{r}^i t_{r{,2}}^i\) and so on. This ensures that each reply remains contextually connected to its preceding sub-replies, preserving the logical progression of the discussion.  }
 
{However, since SemEval and P-Stance do not contain threaded conversations, this step is excluded, and each reply post is processed independently.  
By implementing these preprocessing techniques, we refine the dataset into a structured format, ensuring clarity, consistency, and robustness for stance detection. }  
    
\subsubsection{\textit{Comparison Methods}}
In this section, we discuss the different methods used for comparison and effectiveness of the proposed method is evaluated against different state-of-the-art methods. This evaluation involves a comprehensive examination of state-of-the-art methods for stance classification, ensuring a rigorous and comprehensive analysis. 
We validated our results by reproducing methods and conducting a thorough comparison with state-of-the-art methods on the same datasets.
\begin{enumerate}
\item{Mistral} \cite{jiang2023mistral}
is designed to provide high performance for a variety of NLP tasks. The model supports advanced features such as sliding window attention, making it capable of handling extended contexts for enhanced accuracy in text generation and understanding.
\item{Generative Pre-trained Transformer (GPT-3.5)} \cite{openai2023gpt35turbo}, developed by OpenAI, that utilizes deep learning techniques to understand and generate natural language and code. It features improved contextual understanding and generates relevant text across a variety of tasks, including conversation, summarization, translation, data analysis, gaming narratives, research support, and social media management.
\item{Large Language Model Meta AI (LLaMa3)} \cite{dubey2024llama} is developed by Meta known for its enhanced performance and efficiency. It features advanced capabilities in natural language understanding, improved context retention, and support for various tasks like text generation, summarization, and translation. Applications include chatbots, content generation, and automated customer support, making it versatile for both individual and enterprise use.
\item StanceBERTa\footnote{\href{https://huggingface.co/eevvgg/StanceBERTa}{https://huggingface.co/eevvgg/StanceBERTa}\label{note111}}
 is a fine-tuned version of distilroberta-base model to predict 3 categories of stance (negative, positive, neutral) expressed in a text towards a specific target. StanceBERTa leverages pre-trained language representations fine-tuned on stance detection datasets, which enables the model to understand linguistic patterns and contextual clues pertinent to stance. Also suitable for fine-tuning on hate or offensive language detection.
 \item Bidirectional Encoder Representations from Transformers (BERT) \cite{devlin2018bert}  has been pre-trained in the English language. By utilizing a dataset of labelled phrases, one can train a standard classifier with features generated by BERT as input. This model captures the underlying representation of English language, which can subsequently be leveraged to extract features that are beneficial for downstream tasks.
  \item Robustly optimized BERT Approach (RoBERTa) \cite{liu2019roberta} is an advanced transformer model that builds on BERT's architecture with key modifications to improve performance. RoBERTa is pre-trained on a more extensive and diverse corpus, making it more robust and effective for various NLP tasks. By optimizing BERT's training process and increasing data exposure, RoBERTa achieves state-of-the-art results on numerous benchmarks, such as sentiment analysis, text classification, and question answering.
  \item { Finetuned Language Net (FLAN-T5)} \cite{FLAN-T5}, an advanced version of Text-To-Text Transfer Transformer (T5) \cite{t5}, fine-tuned using the FLAN (Few-Shot Learning with Auxiliary Networks) methodology. It is designed to enhance performance on a wide range of natural language understanding tasks by leveraging few-shot learning techniques. This model improves its ability to handle diverse text generation and comprehension tasks through this fine-tuning approach.
\item Decoding-enhanced BERT with Disentangled Attention (DeBERTa) \cite{he2021debertav3} is a transformer-based model that builds upon BERT and RoBERTa by enhancing the attention mechanism. It introduces disentangled attention to capture token relationships by separating content and positional information, improving contextual understanding. The model also employs an enhanced decoding mechanism that refines the representation of text. 
{  \item Collaborative rOle-infused LLM-based Agents (COLA) \cite{Lan_Gao_Jin_Li_2024}
assigns distinct roles to large language models (LLMs), creating a collaborative system. The LLMs are configured to act as a linguistic expert, a domain specialist, and a social media veteran to provide a multifaceted analysis of the text. Then a final decision-making agent consolidates these prior insights to determine the stance.}
  \item Topic-Agnostic and Topic-Aware Embeddings (TATA) \cite{tata} is a framework designed for stance detection that leverages both general and topic-specific information. It utilizes topic-agnostic embeddings to capture broad contextual features applicable across various topics. By integrating these two types of embedding vectors, TATA enhances the method's ability to detect stances more accurately, regardless of the topic's specificity.
  \item Enhancing Zero-Shot Stance Detection with Contrastive and Prompt Learning (EZSD-CP) \cite{e26040325} leverages contrastive learning and prompt-based techniques to improve model performance without requiring labeled training data.
It presents a gating mechanism that dynamically adjusts the influence of the gate on prompt tokens based on semantic features, improving the relevance between instances and prompts.
  \item Zero-Shot Stance Detection (ZSSD) \cite{liang2022zero} technique is based on the idea that stance detection can be performed without any task-dependent training data. This method employs contrastive learning objectives by learning how to differentiate between relevant and irrelevant stance-based information. This approach handles the challenge of zero-shot learning, allowing for effective stance detection across diverse contexts without extensive labelled datasets.  
 
  \item Dual Contrastive Learning (Dual CL) \cite{chen2022dual} makes use of data augmentation with the help of labels to increase the performance of the method. This approach introduces two contrastive learning objectives: one designed to enhance data of the same class and the other to separate different classes. These objectives help the method to more effectively differentiate texts. The combination of contrastive learning with targeted data augmentation is a proven effective strategy for text classification.
  
  \item Joint Contrastive Learning (Joint CL) \cite{liang2022jointcl} utilizes contrastive learning to align textual representations with the information related to the target. In this way, the effectiveness of the approach is enhanced even for targeted and domain conditions where no additional learning for that task is available.
  
  \item RoBERTa+MLP \cite{prakash-tayyar-madabushi-2020-incorporating} leverages contextual embeddings from RoBERTa, which are concatenated with count-based features and passed through a multi-layer perceptron for final stance classification. By merging these features with pre-trained embeddings, the model benefits from both contextual understanding and quantitative information.
  \item ZeroStance \cite{zhao-etal-2024-zerostance} proposes a novel method for open-domain stance detection using a synthetic dataset called CHATStance, generated through ChatGPT. By providing a task description in the form of a prompt, ZeroStance effectively constructs a cost-efficient and data-efficient dataset for training. Trained on an open-domain model on the synthetic
dataset after proper data filtering indicates that the model, when trained
on this synthetic dataset, shows superior
generalization to unseen targets of diverse
domains. 
{\item Dynamic Experienced Expert Modeling for Stance Detection (DEEM) \cite{wang-etal-2024-deem} introduces a novel framework that leverages dynamic expert modeling to improve stance detection across diverse and evolving topics. Unlike traditional approaches that rely on static models, DEEM employs a dynamic ensemble of ``experienced experts,'' specialized sub-models trained on different domains, which are adaptively weighted based on the input target.}
{\item LLM-Driven Knowledge Injection Advances Zero-Shot and Cross-Target
Stance Detection (LKI-BART) \cite{zhang-etal-2024-llm-driven} proposes innovative techniques to leverage LLMs for knowledge augmentation, enabling models to generalize better to unseen topics or domains. The study demonstrates that LLM-driven knowledge injection significantly outperforms traditional stance detection methods, offering improved accuracy and robustness in identifying stances across diverse and previously unencountered targets.} 
  
\end{enumerate}
{Table \ref{tab:keydiff} presents a comparative analysis of SPLAENet against several state-of-the-art (SOTA) models, highlighting key components such as attention mechanisms, affective features (emotion and sentiment), label fusion, and distance metric learning (DML). SPLAENet introduces several unique advantages that distinguish it from previous approaches in stance detection.
While Hans \textit{ et al.} \cite{tata} employ scaled dot-product attention, their model does not explicitly model bidirectional inter and intra-relationships between source and reply texts.
SPLAENet overcomes this limitation through a dual cross-attention mechanism that captures bidirectional dependencies between source and reply text, enabling a deeper contextual understanding across texts.
Chen\textit{ et al.} \cite{chen2022dual} whose label-aware method is confined to single-text sentiment classification and lacks cross-text modeling, SPLAENet introduces a label-aware fusion mechanism tailored for stance detection in conversational settings, allowing for better alignment between learned features and stance categories. While LKI-BART \cite{zhang-etal-2024-llm-driven} incorporates LLM-generated insights, including latent emotional and rhetorical signals, this method does not explicitly model label emotions.  SPLAENet goes beyond this by modeling emotional alignment between source and reply texts.
Lastly, SPLAENet incorporates distance metric learning (DML) to enhance the discriminability of stance representations. By enforcing proximity between embeddings of similar stances and distancing those of dissimilar stances, the model improves its ability to generalize across varied contexts.
Together, these approaches make SPLAENet a systematic, emotion-aware, and label-sensitive stance detection model, offering a more robust framework for handling the complexity of misinformative social media content.}
\begin{table}[h]
    \centering
    \renewcommand{\arraystretch}{1.2} 
    \begin{tabular}{l c c c c c c c} 
        \toprule
         & Attention & Affective Features & Label Fusion & DML \\
        \midrule
        COLA\cite {Lan_Gao_Jin_Li_2024}  & $\textcolor{red}{\xmark}$ & $\textcolor{red}{\xmark}$ & $\textcolor{red}{\xmark}$ & $\textcolor{red}{\xmark}$ & \\
        TATA \cite{tata} & $\textcolor{blue}{\cmark}$ & $\textcolor{red}{\xmark}$ & $\textcolor{red}{\xmark}$ & $\textcolor{red}{\xmark}$ &  \\
        EZSD-CP \cite{e26040325} & $\textcolor{red}{\xmark}$ & $\textcolor{red}{\xmark}$ & $\textcolor{red}{\xmark}$ & $\textcolor{red}{\xmark}$ &  \\
        ZSSD \cite{liang2022zero}  & $\textcolor{red}{\xmark}$ & $\textcolor{red}{\xmark}$ & $\textcolor{red}{\xmark}$ & $\textcolor{red}{\xmark}$ &   \\
        Dual CL \cite{chen2022dual} & $\textcolor{red}{\xmark}$ & $\textcolor{red}{\xmark}$ & $\textcolor{blue}{\cmark}$ & $\textcolor{blue}{\cmark}$ &  \\
        Joint CL \cite{liang2022jointcl} & $\textcolor{red}{\xmark}$ & $\textcolor{red}{\xmark}$ & $\textcolor{red}{\xmark}$ & $\textcolor{red}{\xmark}$ &   \\
        RoBERTa+MLP \cite{prakash-tayyar-madabushi-2020-incorporating} & $\textcolor{red}{\xmark}$ & $\textcolor{red}{\xmark}$ & $\textcolor{red}{\xmark}$ & $\textcolor{red}{\xmark}$ & \\ 
        ZeroStance \cite{zhao-etal-2024-zerostance}  & $\textcolor{red}{\xmark}$ & $\textcolor{red}{\xmark}$ & $\textcolor{red}{\xmark}$ & $\textcolor{red}{\xmark}$ & \\
        DEEM \cite{wang-etal-2024-deem}  & $\textcolor{red}{\xmark}$ & $\textcolor{red}{\xmark}$ & $\textcolor{red}{\xmark}$ & $\textcolor{red}{\xmark}$ & \\
        LKI-BART \cite{zhang-etal-2024-llm-driven}  & $\textcolor{red}{\xmark}$ & $\textcolor{blue}{\cmark}$ & $\textcolor{red}{\xmark}$ & $\textcolor{red}{\xmark}$ & \\
        \midrule
        \textbf{SPLAENet} & $\textcolor{blue}{\cmark}$ & $\textcolor{blue}{\cmark}$ & $\textcolor{blue}{\cmark}$ & $\textcolor{blue}{\cmark}$ \\

        \bottomrule
    \end{tabular}
    \caption{{Key Differences of SPLAENet with other SOTA Models}}
    \label{tab:keydiff}
\end{table}
\subsubsection{\textit{Evaluation Metrics}}
Stance detection is primarily framed as a classification problem,  where content generated on social media is categorized into various classes, namely, Support (S), Query (Q), Deny (D) and Comment (C). We employ four evaluation metrics, Accuracy(A), Precision(P), Recall(R) and F1-score(F1). To assess the method's performance across all classes, we adopt the macro average for Precision ($\textit{P}_m$), Recall ($\textit{R}_m$), and F1-score ($F1_m$).

Precision for a specific label, where $\textit{label} \in \{\textit{Support (S)}, \textit{Query (Q)}, \textit{Deny (D)}, \textit{Comment (C)}\}$ is defined as the ratio of correctly predicted instances of that label to the total number of instances predicted as that label, as shown in Equation (\ref{eq:P}).
The macro-averaged precision is calculated in Equation (\ref{eq:m_P}):
\begin{subequations}
\begin{equation}
\textit{P} = \frac{\textit{True\_Positive}_{\textit{label}}}{\textit{True\_Positive}_{\textit{label}} + \textit{False\_Positive}_{\textit{label}}}
\label{eq:P}
\end{equation}

\begin{equation}
\textit{P}_m = \frac{1}{N} \sum_{\textit{label} \in \{ S, Q, D, C\}} \frac{\textit{True\_Positive}_{\textit{label}}}{\textit{Total\_Predicted}_{\textit{label}}}
\label{eq:m_P}
\end{equation}
\end{subequations}

Recall assesses the model's ability to correctly identify all instances of a particular class. ${\textit{True\_Positive}_{label}}$ are the correctly predicted instances of the class, while ${\textit{False\_Negative}_{label}}$ are the actual instances of the class that the method failed to identify. This metric is crucial for evaluating method performance, especially in cases with imbalanced class distributions. The formula for recall is shown in Equation (\ref{eq:R}). The macro-averaged recall is calculated in Equation (\ref{eq:m_R}).
\begin{subequations}
 \begin{equation}
\textit { R }=\frac{\textit{True\_Positive}_{label}}{\textit{True\_Positive}_{label}+\textit{False\_Negative}_{label}}
\label{eq:R}
\end{equation}

\begin{equation}
\textit{R}_m = \frac{1}{N} \sum_{\textit{label}  \in \textit \{S,Q,D,C\}} \frac{\textit{True\_Positive}_{\textit{label}}}{\textit{True\_Positive}_{\textit{label}}+\textit{False\_Negative}_{\textit{label}}}
\label{eq:m_R}
\end{equation}
\end{subequations}
F1-score combines precision and recall as their harmonic mean. It is useful for balancing both metrics, particularly when the class distribution is imbalanced, as shown in Equation (\ref{eq:F1}). 
The Macro-F1 score is an averaging method for F1 scores across different classes in a classification problem. It is calculated by taking the simple average of F1-scores for each class, treating each class equally irrespective of the number of instances in each class. The Macro-averaged F1 is calculated in Equation (\ref{eq:m_F}).
\begin{subequations}
\begin{equation}
\textit{F1}=\frac{2 \times \textit { P}_\textit{label} \times \textit { R}_\textit{label}}{\textit { P}_\textit{label}+\textit { R}_\textit{label}}
\label{eq:F1}
\end{equation}

\begin{equation}
\textit{F1}_m = \frac{1}{N} \sum_{\textit{label }  \in { \textit \{ 
S,Q,D,C \}} } {\textit{F1}_{\textit{label}}}
\label{eq:m_F}
\end{equation}
\end{subequations}
The accuracy metric is defined as the quotient of a total number of correctly predicted posts against the total number of posts in the dataset. The formula for Accuracy is given in Equation (\ref{eq:Acc}).
 \begin{equation}
\textit {A}= \sum_{\textit{label}  \in \textit \{S,Q,D,C\}}\frac{\textit{True\_Positive}_{label}}{\textit{Total\_Posts}}
\label{eq:Acc}
\end{equation}
\subsection {\textit{Experimental Results}} \label{label:experiment_results}

This section presents the experimental evaluation, by comparing the performance of the proposed method with baseline and state-of-the-art methods.
All SOTA approaches referenced in this evaluation are implemented on the datasets, ensuring that our results reflect true performance, as summarized in Table \ref{tab:metrics}.

\begin{table}[ht]
\centering

\caption{Performance comparison of SPLAENet with existing baselines across Datasets. The top-performing result is highlighted in \textbf{bold}, while the second-best is \underline{underlined}.}

\label{tab:metrics}
\resizebox{\textwidth}{!}{
\begin{tabular}{lcccccccccccc}
\toprule
\multirow{2}{*}{\textbf{Methods}} & \multicolumn{4}{c}{\textbf{RumourEval}} & \multicolumn{4}{c}{\textbf{SemEval}} & \multicolumn{4}{c}{\textbf{P-Stance}} \\ 
\cmidrule(lr){2-5} \cmidrule(lr){6-9} \cmidrule(lr){10-13}
 & \textit{A} & $\textit{P}_m$ & $\textit{R}_m$ & $\textit{F1}_m$ & \textit{A} & $\textit{P}_m$ & $\textit{R}_m$ & $\textit{F1}_m$ & \textit{A} & $\textit{P}_m$ & $\textit{R}_m$ & $\textit{F1}_m$ \\ 
\midrule
Mistral \cite{jiang2023mistral} & 54.27 & 35.60 & {47.30} & 36.34 & 68.86 & 70.81 & 71.38 & 68.71& \underline{84.61}
&\underline{85.54}&	\underline{84.25}	&	\underline{84.39} \\ 
GPT-3.5 \cite{openai2023gpt35turbo} & 70.91 & 39.21 & 42.90 & 38.32 & 70.94 & 66.26 & 65.17 & 64.65 & 78.03 & 79.43 & 77.52 & 77.51 \\ 
LLaMa3 \cite{dubey2024llama} & 77.17 & 25.81 & 24.68 & 24.48 & 69.34 & 68.27 & \underline{73.64} & 68.03& 81.83 & 82.05 & 81.62 & 81.70 \\ 
\midrule

StanceBERTa\footnote{{eevvgg/StanceBERTa}} & 81.06 & 30.46 & 25.23 & 23.26 & 63.65 & 57.89 & 55.96 & 56.72 & 47.84&	23.92&	50.00		&32.36 \\ 
BERT \cite{devlin2018bert} & 84.57 & 21.14 & 25.00 & 22.91 & 65.65 & 63.69 & 46.48 & 46.57 & 62.87 & 65.86 & 66.85 & 61.87 \\ 
RoBERTa \cite{liu2019roberta} & 82.97 & 33.28 & 36.68 & \underline{42.69} & 66.93 & 70.15 & 48.50 & 48.45 & 67.88 & 66.87 & 69.86 & 69.81 \\
FLAN-T5 \cite{FLAN-T5} & 84.96 & 36.78 & 32.12 & 33.18 & 60.77 & 52.16 & 47.59 & 48.48 & 77.12 & 73.17 & 72.34 & 70.12 \\ 
DeBERTa \cite{he2021debertav3} & 84.62 & 37.83 & 25.47 & 23.86 & 65.89 & 61.52 & 50.29 & 51.66 & 58.76 & 51.90 & 56.14 & 56.29 \\ 
\midrule

{COLA \cite{Lan_Gao_Jin_Li_2024}} & {63.00} & {37.23 }& {46.56 }& {38.05 }& {60.37 }& {67.13} & {59.88} & {56.31} & {82.15} & {83.41} & {82.89} & {81.82} \\
TATA \cite{tata} & 80.20 & 40.23 & 36.24 & 37.19 & \underline{74.90} & \underline{71.66} & 73.48 & \underline{71.34} & {83.26} & {83.32} & 83.37& {83.26} \\ 
EZSD-CP \cite{e26040325} & 81.04 & 37.73 & 39.90 & 37.63 & 67.57 & 63.99 & 68.45 & 65.28 & 77.75 & 77.71 & 77.75 & 77.72 \\ 
ZSSD \cite{liang2022zero} & 83.36 & 42.71 & 36.01 & 37.12 & 72.02 & 39.87 & 33.42 & 69.73 & 83.12 & 83.14 & 83.03 & 83.07 \\
Dual CL \cite{chen2022dual} & 84.80 & 40.66 & 27.60 & 27.60 & 69.58 & 66.20 & 62.17 & 63.63 & 78.62 & 78.62 & 78.52 & 78.55 \\ 
Joint CL \cite{liang2022jointcl} & - & - & - & - & 71.26 & 67.89 & 72.29 & 69.40 & 69.90 & 70.20 & 71.80 & 72.86 \\
RoBERTa+MLP \cite{prakash-tayyar-madabushi-2020-incorporating} & \underline{85.31} & \underline{46.12} & 42.48 & 41.32 & 72.38 & 68.99 & 72.54 & 69.84 & {83.96}&	{83.96}&	{83.89}&{83.91} \\
ZeroStance \cite{zhao-etal-2024-zerostance} & {-} & {-} &- & - &60.05&	63.34&	64.49	&	58.73 & 77.98 & 81.56 & 78.69 & 77.60 \\
{DEEM \cite{wang-etal-2024-deem}}&{68.85} &{39.49 } & {\textbf{52.73}} &{ 42.38}& {73.90} &{70.20}&	{71.02}&	{70.17}&{82.32}	 &{85.53}  & {80.81} & {83.73 }\\
{LKI-BART  \cite{zhang-etal-2024-llm-driven}}&{74.21} & {42.93}  & {43.14} &{40.24}& {74.21} &{69.12}&	{68.11}&{	60.78}&{83.52}	 &{82.76}  & {79.12} &{ 82.61 }\\
\midrule
\textbf{SPLAENet} & \textbf{86.50} & \textbf{60.38} & \underline{48.33} & \textbf{51.52} & \textbf{75.26} & \textbf{72.23} & \textbf{73.92} & \textbf{72.50} & \textbf{85.67}	&\textbf{85.93}&	\textbf{85.48}	&	\textbf{85.58}\\
\bottomrule
\end{tabular}
}
\end{table}

\subsubsection{\textit{Performance Evaluation on RumourEval Dataset}}
Table \ref{tab:metrics} presents a comparison of results of various stance detection methods on RumourEval \cite{gorrell-etal-2019-semeval} dataset. It demonstrates that SPLAENet achieves an accuracy of 86.50\%, a precision of 60.38\%, a recall of 48.33\%, and an F1-score of 51.52\%.
It shows substantial improvements over Mistral \cite{jiang2023mistral}, surpassing it by 32.23\% in accuracy, 24.78\% in precision, 1.03\% in recall, and 15.18\% in F1-score. Similarly, compared with GPT-3.5 \cite{openai2023gpt35turbo}, and LLaMa3 \cite{dubey2024llama}, our model outperforms all three llm-based methods. Since SPLAENet is trained primarily on the stance detection dataset, allowing it to understand the details better than the general training used by LLMs.  
Among base models, SPLAENet outperforms RoBERTa  \cite{liu2019roberta}, achieving 3.53\% higher accuracy, along with improvements of 27.10\% in precision, 11.65\% in recall, and 8.83\% in F1-score. These gains can be attributed to the absence of a dual cross-attention mechanism between source and reply texts in RoBERTa.
SPLAENet also outperforms StanceBERTa\textsuperscript{\ref{note111}}, 
 BERT \cite{devlin2018bert}, FLAN-T5 \cite{FLAN-T5}, and RoBERTa \cite{he2021debertav3}.
In comparison with other state-of-the-art methods, SPLAENet demonstrates superior performance. Against TATA \cite{tata}, EZSD-CP \cite{e26040325}, ZSSD \cite{liang2022zero}, Dual CL \cite{chen2022dual}, RoBERTa+MLP \cite{prakash-tayyar-madabushi-2020-incorporating}, and ZeroStance \cite{zhao-etal-2024-zerostance}, our method reaffirms its effectiveness and robustness in stance detection. {Compared to LLM-based approaches, our method outperforms COLA \cite{Lan_Gao_Jin_Li_2024} by substantial margins, 23.50\% accuracy, 23.15\% precision, 1.77\% recall, and 13.47\% F1-score. SPLAENet also surpasses LKI-BART \cite{zhang-etal-2024-llm-driven} across all key metrics by 12.29\% in accuracy, 17.45\% in precision, 5.19\% in recall, and 11.28\% in F1-score. While DEEM \cite{wang-etal-2024-deem} demonstrates a 4.40\% gain in recall, our model achieves superior performance in accuracy, precision, and F1-score.} These advancements stem from SPLAENet's innovative use of attention mechanism, emotion sysnthesis, and integration of label information to model the semantic proximity between labels, source text, and reply text.
\subsubsection{\textit{Performance Evaluation on SemEval Dataset}}
We compare the results of our method on SemEval \cite{mohammad2016semeval} dataset, against previously mentioned techniques. Table \ref{tab:metrics} demonstrates that our method achieves an accuracy of 75.26\%, a precision of 72.23\%, a recall of 73.92\%, and a F1-score of 72.50\%.
Among large language models, SPLAENet outperforms GPT-3.5 \cite{openai2023gpt35turbo}, achieving a notable 4.32\% improvement in accuracy, along with consistent gains of 5.97\%, 8.75\%, and 7.85\% in precision, recall, and F1-score, respectively. This is attributed to the lack of domain-specific knowledge and handcrafted features in GPT-3.5. Similarly, SPLAENet surpasses Mistral \cite{jiang2023mistral} and LLaMa3 \cite{dubey2024llama}, further demonstrating its superiority in task-specific performance.
 
Comparing the baseline models, 
StanceBERTa\textsuperscript{\ref{note111}}, primarily fine-tuned for stance detection, performs relatively less, with an accuracy of 63.65\% and an F1-score of 56.72\%, well below SPLAENet's performance.
Furthermore, SPLAENet is also compared with RoBERTa \cite{liu2019roberta}, BERT \cite{devlin2018bert},  DeBERTa \cite{he2021debertav3}, and FLAN-T5 \cite{FLAN-T5}, highlighting SPLAENet’s effectiveness in this domain.
{In comparison to state-of-the-art methods, SPLAENet consistently outperforms COLA \cite{Lan_Gao_Jin_Li_2024}, achieving a 14.89\% improvement in accuracy and showing gains across all other metrics. These improvements are attributed to SPLAENet’s integration of a dual cross-attention mechanism, emotion synthesis, and distance metric learning, which collectively enhance the model’s ability to quantify and align the relationships between source and reply text features.} Although TATA \cite{tata} demonstrates strong performance, SPLAENet leverages emotion alignment between source and reply text features to achieve an overall improvement of  0.36\% in accuracy, 1.16\% in F1-score, and 0.44\% in recall. When compared to EZSD-CP \cite{e26040325}, SPLAENet shows substantial improvements of 7.69\% in accuracy, 7.22\% in F1, and 5.47\% in recall. SPLAENet also shows superior performance over methods such as ZSSD \cite{liang2022zero}, and Dual CL \cite{chen2022dual}. Lastly, SPLAENet outperforms RoBERTa+MLP \cite{prakash-tayyar-madabushi-2020-incorporating}, with gains of 2.88\%, 3.24\%, 1.38\%, and 2.66\% in the respective metrics since it lacks cross-attention module to capture inter and intra-relationship.
\subsubsection{\textit{Performance Evaluation on P-Stance Dataset}}
As seen in the Table \ref{tab:ablation-pstance}, SPLAENet outperforms existing state-of-the-art methods with a significant margin. Compared to Mistral \cite{jiang2023mistral} performs with an accuracy of 84.61\%, precision of 85.54\%, recall of 84.25\%, and an F1-score of 84.39\%, falling short by 1.06\%, 0.39\%, 1.23\%, and 1.19\% when compared to SPLAENet.
Similarly, TATA \cite{tata}, which has an accuracy of 83.26\%, precision of 83.32\%, recall of 83.37\%, and F1 of 83.26\%, SPLAENet shows a performance gain of 2.41\%, 2.61\%, 2.11\%, and 2.32\%, respectively.
Also,  ZeroStance \cite{zhao-etal-2024-zerostance}, with an F1-score of 77.60\%, precision of 81.56\%, and recall of 78.69\%, has a significant performance gap of 7.98\%, 4.37\%, and 6.79\%, respectively, against SPLAENet.
Finally, RoBERTa+MLP \cite{prakash-tayyar-madabushi-2020-incorporating}, while decent, lags with an accuracy of 83.96\% and an F1-score of 83.91\%, showing a difference of 1.71\% and 1.67\% compared to SPLAENet. Our method significantly outperforms DEEM \cite{wang-etal-2024-deem}, achieving 3.35\% higher accuracy, 0.40\% higher precision, 4.67\% higher recall, and 1.85\% higher F1-score. These comparisons highlight SPLAENet’s superior performance, demonstrating its ability to effectively handle stance detection tasks with notable improvements in accuracy, precision, recall, and F1-score.

\begin{figure}[h!]
    \centering    
    \hspace{1cm}    \includegraphics[width=\textwidth]{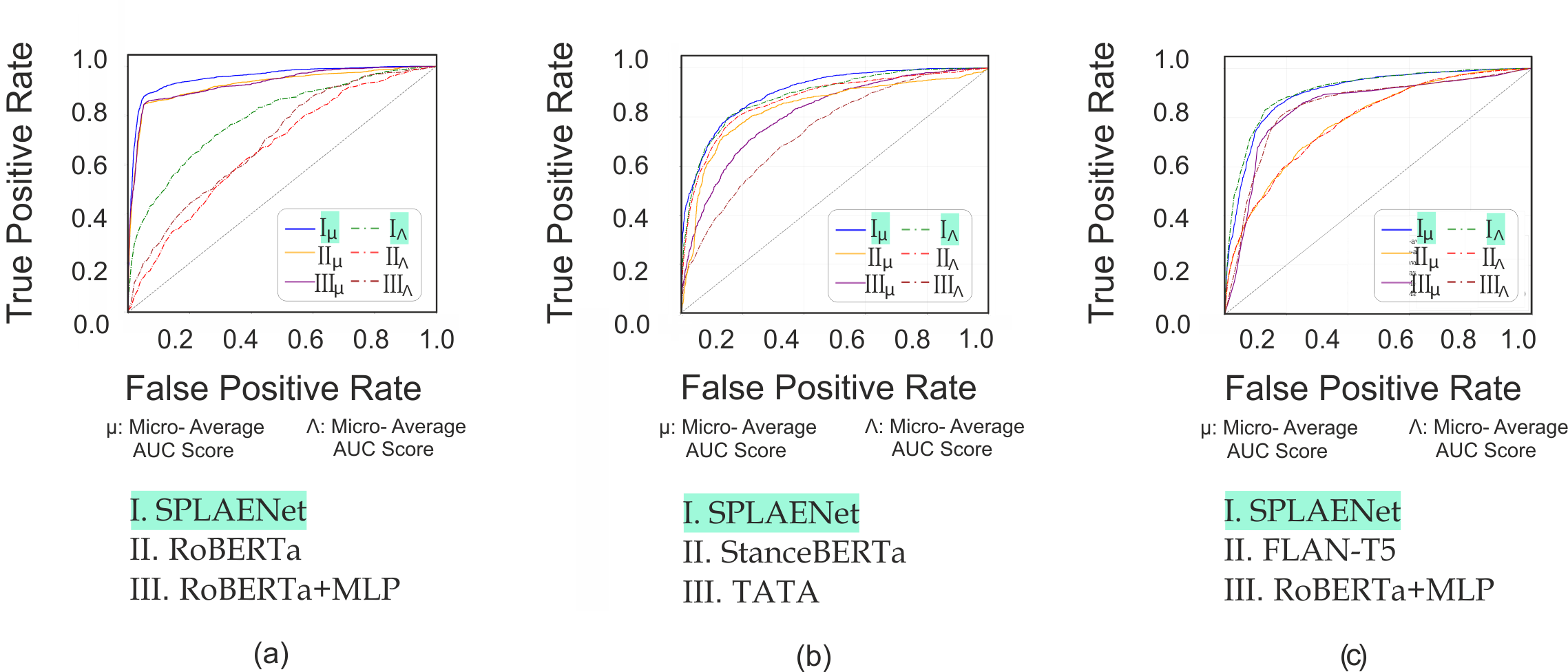}
    \caption{Analysis of ROC curves on Datasets (a) RumorEval (b) SemEval (c) P-Stance}
      \label{fig:ROC_CURVE}
\end{figure}

Figure \ref{fig:ROC_CURVE} presents the receiver operating characteristic (ROC) curves, comparing the performance of the proposed method with top-performing base models and state-of-the-art baseline methods across all three datasets: \ref{fig:ROC_CURVE}(a) RumourEval \cite{gorrell-etal-2019-semeval}, \ref{fig:ROC_CURVE}(b) SemEval \cite{mohammad2016semeval}, and \ref{fig:ROC_CURVE}(c) P-Stance \cite{p-stance}.
\subsection{Ablation Study}
In this section, we analyse the contribution of individual components to the performance of our stance detection method through ablation studies on all three datasets.
\subsubsection{\textit{Ablation Analysis on RumourEval Dataset}}
 The ablation results of the experiments are summarised in Table \ref{tab:ablation-rumour}.
Figure \ref{fig:bargraph-rumour}(a) presents a comparison of the results obtained from different attention mechanisms, while \ref{fig:bargraph-rumour}(b) illustrates the evaluation of various features applied to the RumourEval \cite{gorrell-etal-2019-semeval} dataset.
In the following section, we explore the importance of attention and features in SPLAENet.
\begin{table}[h!]
\centering
\resizebox{\textwidth}{!}{
\begin{tabular}{lcccccccccccc}
\toprule
\textbf{Component} & \textbf{Methods} & \textbf{$A$(\%)} &   \textbf{{$P_m$}(\%)} &  \textbf{{$R_m$}(\%)} &  \textbf{{$F1_m$}(\%)} \\
\midrule
\multirow{3}{*}{Attention Mechanism} 
    & SPLAENet w/o DCA & {84.68} & 51.15 & 43.15 & 44.13 \\
    & SPLAENet w/o HAN & 85.88 & 55.65 & 42.74 & 43.89 \\
    & SPLAENet w/o Both & 84.85 & {33.85} & {35.58} & {34.58} \\
\midrule
     \multirow{3}{*}{Feature Combination} 
    & SPLAENet w/o Label Fusion & 85.89 & 58.72 & {43.97} & {46.45} \\
    & SPLAENet w/o Emotion & 86.05 & 59.29 & 46.34 & 48.85 \\ 
    & SPLAENet w/o Feature Closeness & {84.39} &{53.46} & 46.85 & 48.86 \\ 
\midrule
\multicolumn{2}{l}{\textbf{SPLAENet}} & \textbf{86.50} & \textbf{60.38} & \textbf{48.33} & \textbf{51.52} \\       
\bottomrule
\end{tabular}
}
\caption{Evaluation of SPLAENet's performance with Feature Combinations and Attention Mechanisms on the RumourEval dataset}
\label{tab:ablation-rumour}
\end{table}

\begin{enumerate}[label=\alph*)]
  \item{\textit{Ablation on Attention Mechanisms}: In this section, we present the
experimental analysis of SPLAENet with different
attention mechanisms. SPLAENet w/o DCA, SPLAENet w/o HAN, and
SPLAENet w/o Both are variants of our proposed method by
eliminating dual cross-attention (DCA) and hierarchical attention networks (HAN) respectively. 
To demonstrate the usability of attention mechanism in Table \ref{tab:ablation-rumour}, we removed it while retaining the remaining features.
First, we analyzed the variant excluding the dual cross-attention mechanism, referred to as SPLAENet w/o DCA. The results indicated that integrating DCA mechanism leads to a 1.82\% increase in accuracy compared to the proposed method. Additionally, the presence of DCA improved precision by 9.23\%, recall by 5.18\%, and F1-score by 7.39\%. These enhancements emphasize the significant performance benefits derived from DCA. Next, we examined the variant excluding the hierarchical attention network, known as SPLAENet w/o HAN, which also showed superior performance in the proposed method, with a 0.62\% increase in accuracy. The inclusion of the HAN contributed to a precision increase of 4.73\%, a recall increase of 5.59\%, and a F1-score increase of 7.63\%. This underlines the critical role played by HAN in capturing and utilizing hierarchical relationships within data.
Lastly, SPLAENet w/o Both variant, which excludes both DCA and HAN mechanisms, illustrates the most pronounced impact on performance. The presence of these mechanisms results in an accuracy that was increased by 1.65\%. However, when both mechanisms are utilized, the method achieved a remarkable increment, including precision improved by 26.49\%, recall improved by 12.75\%, and F1-score improved by 16.94\%. These substantial enhancements reinforce the critical contributions of DCA and HAN mechanisms in boosting the performance of SPLAENet. }
\begin{figure}[h!]
    \centering    
    \hspace{1cm}    \includegraphics[width=0.93\textwidth]{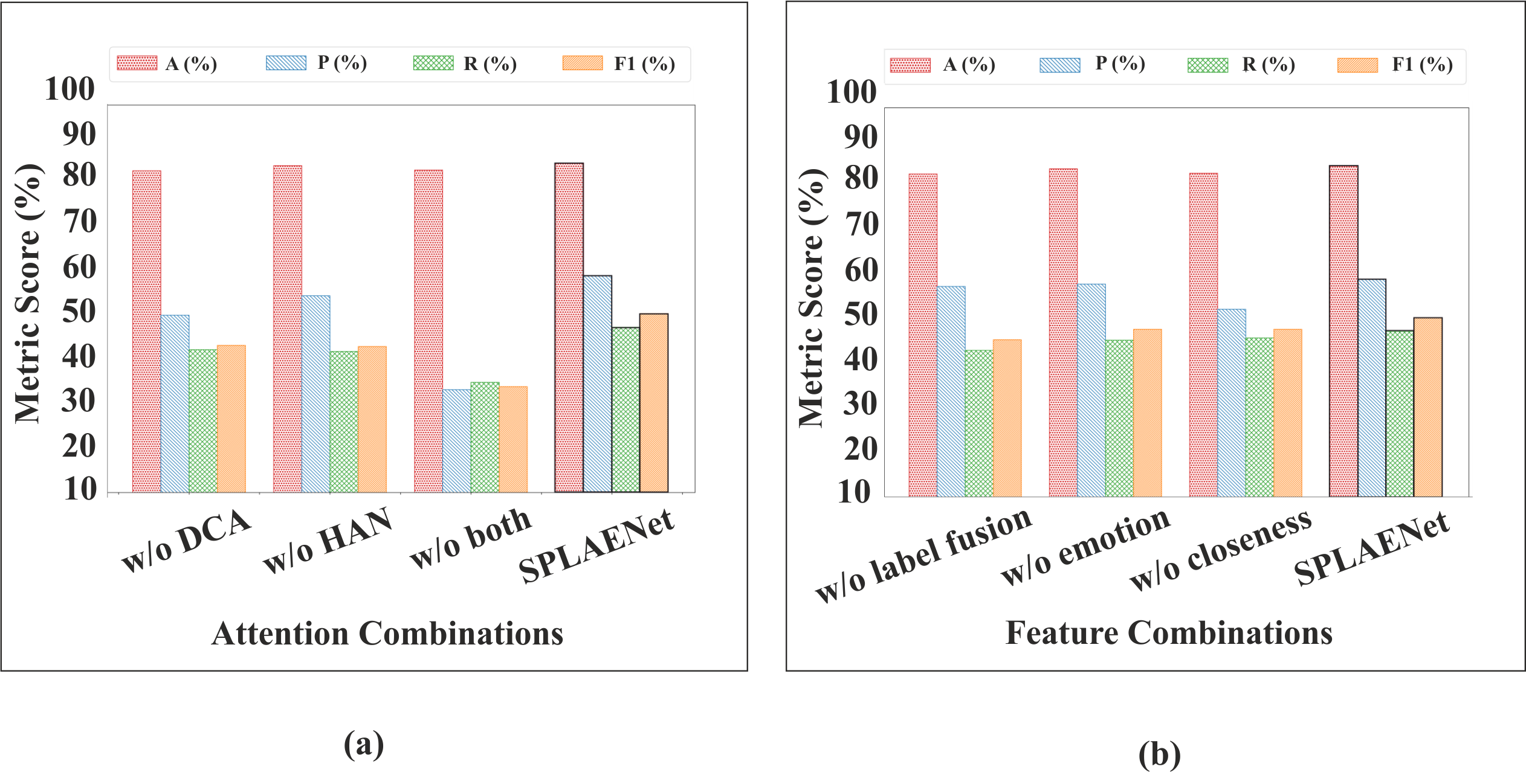}
    \caption{Ablation of (a) Attention and (b) Feature Importance on RumourEval Dataset}
      \label{fig:bargraph-rumour}
\end{figure}
\item{\textit{Ablation on Features Importance}:
In this section, we provide the experimental analysis of SPLAENet with various features. We evaluate three variants of our method namely SPLAENet w/o Label Fusion, SPLAENet w/o Emotion, and
SPLAENet w/o Feature Closeness which excludes
label fusion, emotions, and feature closeness between embedding vector of source and reply text respectively.
To highlight the significance of these features we excluded each feature individually while keeping the others intact. Table \ref{tab:ablation-rumour} presented an experimental analysis of SPLAENet with different features to evaluate their impact on the method’s performance.
The variant labeled SPLAENet w/o Label Fusion, exhibited significant improvements when label fusion was included. Accuracy is increased by 0.61\%, precision by 1.66\%, recall by 4.36\%, and F1-score by 5.07\%, suggesting that label fusion plays a crucial role in balancing precision and recall effectively. Similarly, the variant SPLAENet w/o Emotion, which omitted the emotion feature, demonstrated benefits when the emotion component was included, leading to increases of 0.45\% in accuracy, 1.09\% in precision, 1.99\% in recall, and 2.67\% in F1-score. The balanced improvement across all performance metrics, particularly in the F1-score, reinforces the emotional component's value in enhancing the method's performance.
The SPLAENet w/o Feature Closeness variant, which excluded the feature closeness component, showed that including this feature led to a more substantial increase in performance. Accuracy is improved by 2.11\%, precision by 6.92\%, recall by 1.48\%, and F1-score by 2.66\%.
The inclusion of DCA, HAN, label fusion, emotion, and feature closeness all led to considerable improvements in performance, with attention and feature closeness showing a pronounced effect on all evaluation metrics.}
\end{enumerate}
\subsubsection{\textit{Ablation Analysis on SemEval Dataset}}
In this section, we present the experimental analysis of SPLAENet, incorporating different attention mechanisms and features performed on SemEval \cite{mohammad2016semeval} dataset. The results of these experiments are summarized in Tables \ref{tab:ablation-sem}. Figure \ref{fig:bargraph-sem}(a) demonstrates a comparison graph between the results obtained from attention mechanisms, while \ref{fig:bargraph-sem}(b) examines features for the SemEval dataset.

\begin{table}[h!]
\centering
\resizebox{\textwidth}{!}{
\begin{tabular}{lcccccccccccc}
\toprule
\textbf{Component} & \textbf{Methods} & {$A$(\%)} &   \textbf{{$P_m$}(\%)} &  \textbf{{$R_m$}(\%)} &  \textbf{{$F1_m$}(\%)} \\
\midrule
\multirow{3}{*}{Attention Mechanism} 
    & SPLAENet w/o DCA & 72.13 & 69.20 & 70.53 & 68.94 \\
    & SPLAENet w/o HAN & 72.93 & {69.17}& {70.15} & 69.06 \\
& SPLAENet w/o Both & {69.01} & {65.42} & {69.99} & {66.80} \\ 
\midrule
\multirow{3}{*}{Feature Combination} 
& SPLAENet w/o Label Fusion & {71.66} & {67.71} & {69.26} & {67.67} \\
    & SPLAENet w/o Emotion & 73.41 & 70.34 & 71.85 & 70.29 \\
    & SPLAENet w/o Feature Closeness & 74.05 & 70.46 & 72.83 & 71.15 \\
\midrule
\multicolumn{2}{l}{\textbf{SPLAENet}} & \textbf{75.26} & \textbf{72.23} & \textbf{73.92} & \textbf{72.50} \\       
\bottomrule
\end{tabular}
}
\caption{Evaluation of SPLAENet's Performance with Feature Combinations and Attention Mechanisms on the SemEval-2019 Dataset}
\label{tab:ablation-sem}
\end{table}

\begin{enumerate}[label=\alph*)]
\item{\textit{Ablation on Attention Mechanisms}:
In Table \ref{tab:ablation-sem}, we first examine the performance of the variant SPLAENet w/o DCA, which achieves an accuracy of 72.13\%. This performance highlights an increase of 3.13\% for the SPLAENet method's accuracy. The precision sees an increase of 3.03\%, recall shows an increase of 3.39\%, and F1-score increases by 3.56\%. These improvements demonstrate the substantial impact of DCA on overall method performance.
The performance of SPLAENet w/o HAN shows an accuracy of 72.93\%, reflecting an increase of 2.33\% in accuracy compared to the proposed method. The precision indicates an increase of 3.06\%, while recall shows an increase of 3.77\%, and F1-score increases by 3.44\%. These enhancements signify the importance of HAN in boosting the method's effectiveness.
When both DCA and HAN are excluded in the variant SPLAENet w/o Both, and compared to the proposed method SPLAENet, this proposed method demonstrates an increase of 6.25\% in accuracy, 6.81\% in precision, 3.93\% in recall, and 5.7\% in F1-score.
The significant decline in performance when both attention mechanisms are removed underscores the combined impact of dual cross-attention and the hierarchical attention network in enhancing the method’s effectiveness.
\begin{figure}[h!]
    \centering
    \hspace{1cm}    \includegraphics[width=0.93\textwidth]{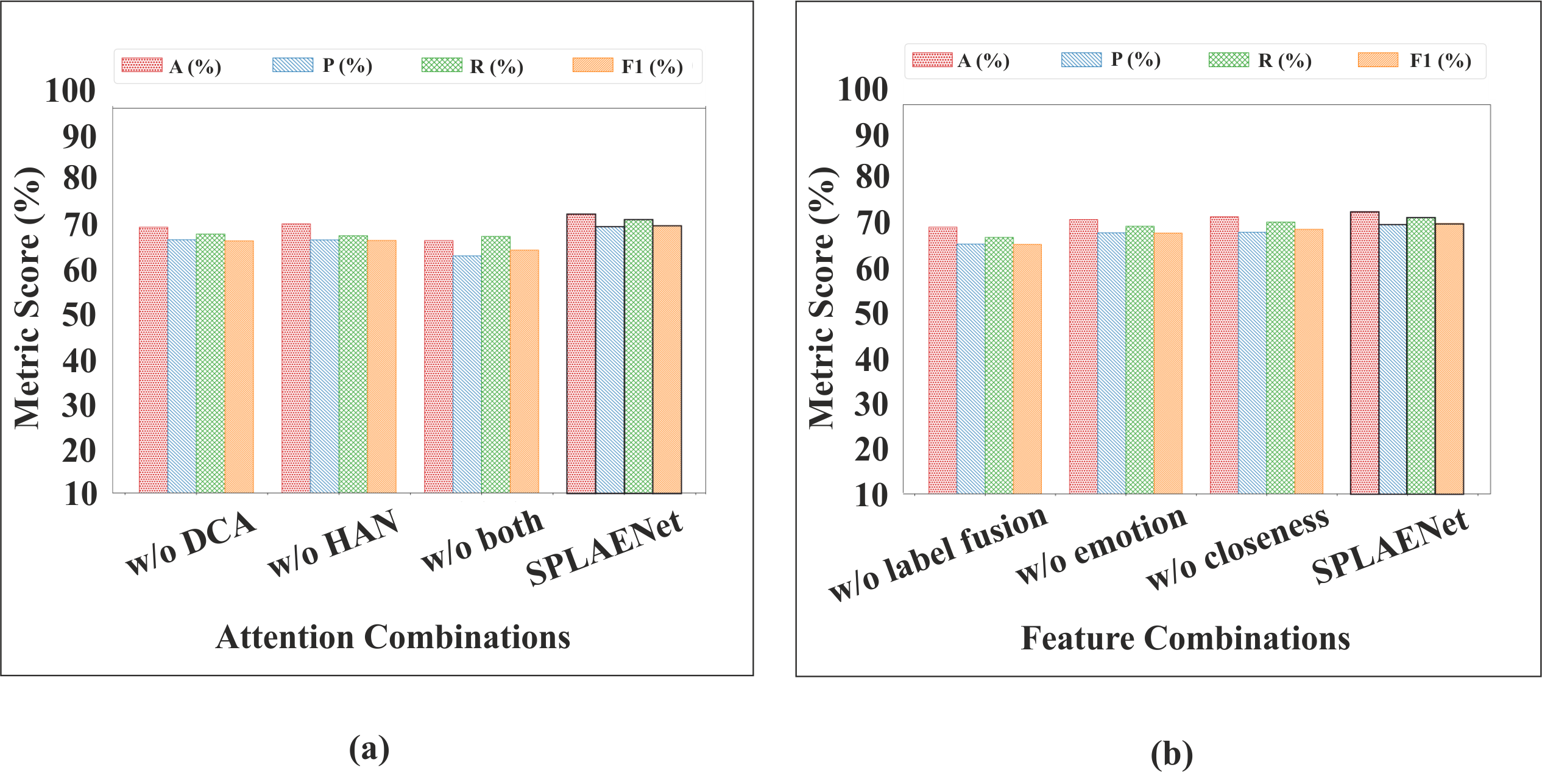} 
    \caption{Ablation of (a) Attention and (b) Feature Importance on the SemEval Dataset}
    \label{fig:bargraph-sem}
\end{figure}
}
\item{\textit{Ablation on Features}: This section evaluates the performance improvements gained by incorporating various features into the SPLAENet method compared to their absence.
The results shown in Table \ref{tab:ablation-sem} demonstrate that when label fusion is added, the method's accuracy improves by 3.60\%, with gains of 4.52\% in precision, 4.66\% in recall, and 4.83\% in the F1-score. This significant improvement demonstrates the crucial role of label fusion in enhancing overall method performance.
Similarly, adding emotion features leads to a performance boost, increasing accuracy by 1.85\%, precision by 1.89\%, recall by 2.07\%, and F1-score by 2.21\%. While the improvement from emotion features is smaller compared to label fusion, these results underscore the importance of emotional context.
This method is further improved by adding feature closeness, which captures the closeness of source text and its reply text. This results in an improvement of 1.21\% in accuracy, 1.77\% in precision, 1.09\% in recall, and 1.35\% in F1-score. This suggests that feature closeness contributes to a better understanding of the proximity between texts, further refining the method's predictive capabilities.
Overall, label fusion, emotion, and feature closeness provide valuable insights and contribute to progressive performance improvements. The proposed SPLAENet method, which includes all these features, achieves the most significant gains in performance, showing that combining all three features leads to the best performance.
}
\end{enumerate}
\subsubsection{\textit{Ablation Analysis on P-Stance Dataset}}
This section presents the experimental analysis of SPLAENet, evaluating its performance with various attention mechanisms and features on the P-Stance \cite{p-stance} dataset. The results of these experiments are detailed in Table \ref{tab:ablation-pstance}. Figure \ref{fig:bargraph-pstance}(a) illustrates a comparative analysis of the results of different attention mechanisms, while Figure \ref{fig:bargraph-pstance}(b) highlights feature-related performance. 
\begin{table}[h!]
\centering
\resizebox{\textwidth}{!}{
\begin{tabular}{lcccccccccccc}
\toprule
\textbf{Component} & \textbf{Methods} & \textbf{{$A$(\%)}} &   \textbf{{$P_m$}(\%)} &  \textbf{{$R_m$}(\%)} &  \textbf{{$F1_m$}(\%)} \\
\midrule
\multirow{3}{*}{Attention Mechanism} 
    & SPLAENet w/o DCA & 84.14 & 84.27 & 83.99 & 84.06 \\
    & SPLAENet w/o HAN & 85.11 & 85.16 & 85.01 & 85.06 \\
    & SPLAENet w/o Both & {83.40} & {83.62}& {83.20} & {83.29} \\ 
\midrule

\multirow{3}{*}{Feature Combination} 
    & SPLAENet w/o Label Fusion & 84.89 & 84.87 & 84.83 & 84.85 \\
    & SPLAENet w/o Emotion & {83.35} & 83.44 & {83.21} & {83.27} \\
    & SPLAENet w/o Feature Closeness & 83.40 &{83.36} & 83.39 & 83.38 \\
\midrule
\multicolumn{2}{l}{\textbf{SPLAENet}} & \textbf{85.67}	&\textbf{85.93}&	\textbf{85.48}		&\textbf{85.58}\\       
\bottomrule
\end{tabular}
}
\caption{Evaluation of SPLAENet's Performance with Feature Combinations and Attention Mechanisms on the P-Stance Dataset}
\label{tab:ablation-pstance}
\end{table}

\begin{enumerate}[label=\alph*)]
\item{\textit{Ablation on Attention Mechanisms}: 
For attention mechanisms, the results in Table \ref{tab:ablation-pstance}, show that excluding dual cross-attention (DCA) mechanism reduces the model's performance, yielding an accuracy of 84.14\%, precision of 84.27\%, recall of 83.99\%, and an F1-score of 84.06\%. Removing the hierarchical attention network (HAN) slightly affects metrics, with an accuracy of 85.11\%, precision of 85.16\%, recall of 85.01\%, and an F1-score of 85.06\%. When both DCA and HAN are excluded, the performance further declines to an accuracy of 83.40\%, precision of 83.62\%, recall of 83.20\%, and an F1-score of 83.29\%. SPLAENet without Both yields the lowest performance across all metrics, with values underlined to indicate the significant decrease compared to other configurations. These findings underscore the significant contributions of both DCA and HAN, individually and synergistically, to the overall performance.

\begin{figure}[h!]
    \centering
    \hspace{1cm}    \includegraphics[width=0.93\textwidth]{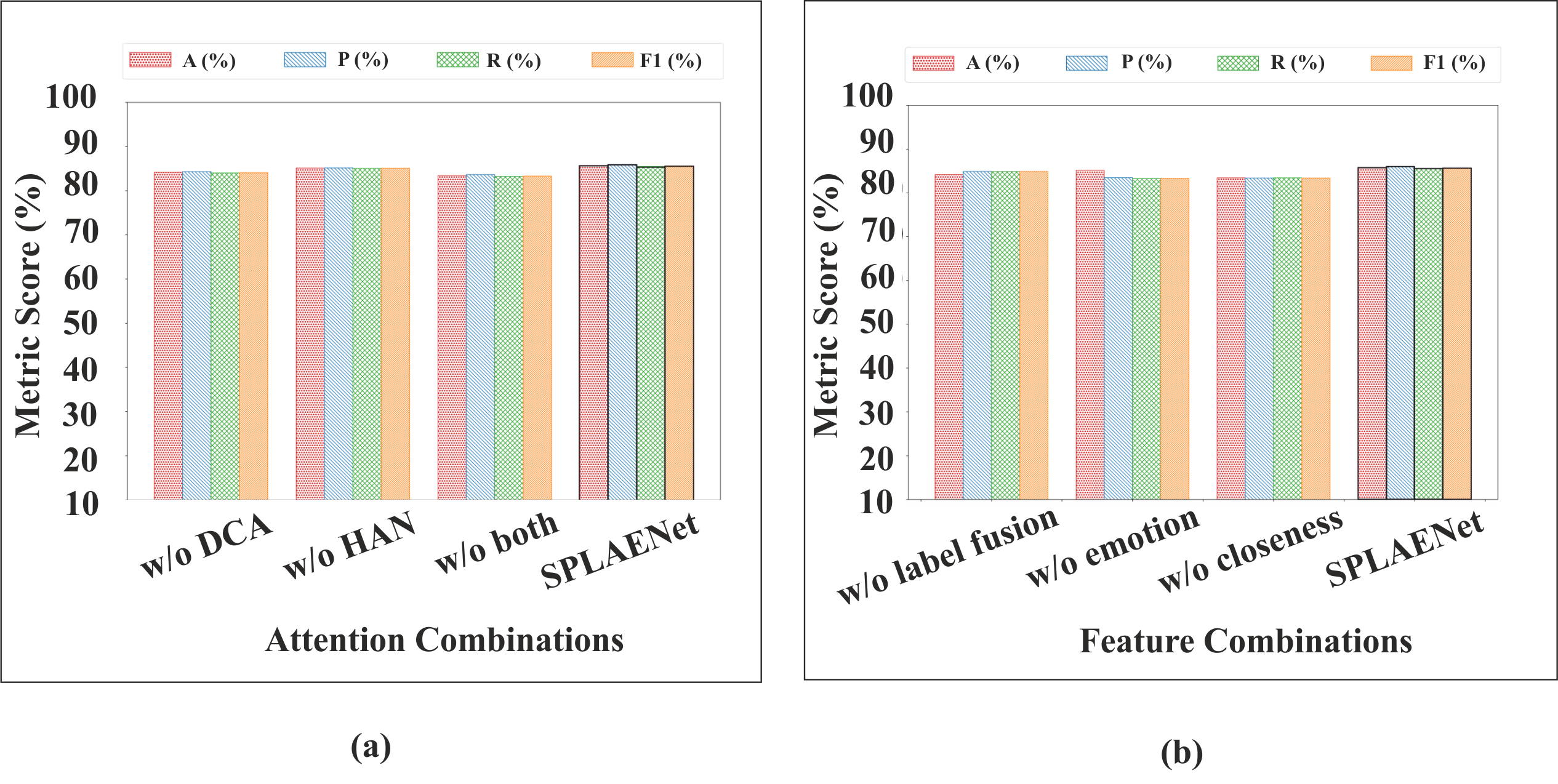} 
    \caption{Ablation of (a) Attention and (b) Feature Importance on the P-Stance Dataset}
    \label{fig:bargraph-pstance}
\end{figure}
}
\item{\textit{Ablation on Features}: 
Regarding feature combinations demonstrated in Table \ref{tab:ablation-pstance}, removing label fusion leads to an accuracy of 84.89\%, precision of 84.87\%, recall of 84.83\%, and an F1-score of 84.85\%, indicating its role in aligning features with stance labels. The exclusion of emotion features has a more pronounced effect, resulting in an accuracy of 83.35\%, precision of 83.44\%, recall of 83.21\%, and an F1-score of 83.27\%, highlighting the importance of incorporating emotional context. Similarly, removing feature closeness affects the results, with an accuracy of 83.40\%, precision of 83.36\%, recall of 83.39\%, and an F1-score of 83.38\%, showing its value in maintaining feature alignment.
This highlights the importance of these components in enhancing the method's stance detection capabilities across various datasets.

}
\end{enumerate}

\subsubsection{\textit{Qualitative Analysis}}
{This section provides a qualitative analysis to assess how various methods perform in identifying stances. In Table \ref{Qualitative Analysis}, we present a set of four user-generated content, comprising source texts, reply texts, associated emotions, and the predicted stance for each post, to examine stances detected by various methods. The example post is taken from RumourEval dataset, where blue symbols indicate the correct stance and red symbols indicate incorrect predictions. Correct stance refers to predictions that align with the ground truth stance label.}
\begin{table*}[!h]
\centering
\resizebox{\textwidth}{!}{\begin{tabular}{lcccc}
\toprule
 & \textbf{Post 1} & \textbf{Post 2} & \textbf{Post 3} & \textbf{Post 4}  \\
\midrule
\textbf{Source Text} & \begin{minipage}[t]{0.16\textwidth}\centering This is crazy \#capetown \#capestorm \#weather \#forecast\end{minipage} & 
\begin{minipage}[t]{0.18\textwidth}\centering 
I bet. You have never seen these rare natural phenomena. Lighting hits a river. What a sight. Incredible Indeed.
\end{minipage} &
\begin{minipage}[t]{0.18\textwidth}\centering 
Jim Bates actually lives in Puerto Rico. They are getting help from Army, Navy, FEMA, Ministers and celebrities.\end{minipage} & 
\begin{minipage}[t]{0.18\textwidth}\centering Is it true that your battery life will improve if you let it charge to 100 before you use it when you first get it?\end{minipage} \\
\\
{\textbf{Emotions}} & \begin{minipage}[t]{0.16\textwidth}\centering  {joy, positive, trust}\end{minipage} & 
\begin{minipage}[t]{0.18\textwidth}\centering 
{-}
\end{minipage} &
\begin{minipage}[t]{0.18\textwidth}\centering {-}
\end{minipage} & 
\begin{minipage}[t]{0.18\textwidth}\centering {joy,  positive, trust}\end{minipage}
 \\\midrule
\textbf{Reply Text} & \begin{minipage}[t]{0.16\textwidth}\centering It is legit, a friend also sent me a video but different angle.\end{minipage} & 
\begin{minipage}[t]{0.18\textwidth}\centering 
 Did it happen in Assam?\end{minipage} &
\begin{minipage}[t]{0.18\textwidth}\centering 
 Is it fake? \end{minipage} & 
\begin{minipage}[t]{0.18\textwidth}\centering Depends on the type.\end{minipage}  \\ \\
{\textbf{Emotions}} & \begin{minipage}[t]{0.16\textwidth}\centering  {joy, positive, trust}\end{minipage} & 
\begin{minipage}[t]{0.18\textwidth}\centering 
{anticipation}
\end{minipage} &
\begin{minipage}[t]{0.18\textwidth}\centering {negative}
\end{minipage} & 
\begin{minipage}[t]{0.18\textwidth}\centering {-}\end{minipage} \\
 \midrule \\
\textbf{Ground Truth} & \textcolor{blue}{Support} & \textcolor{blue}{Query} & \textcolor{blue}{Deny} & \textcolor{blue}{Comment} \\
\midrule
\multirow{2}{*}{\textbf{Methods}} & \multicolumn{4}{c}{\textbf{Predictions}} \\
\cmidrule(lr){2-5}
 & \textbf{Post 1} & \textbf{Post 2} & \textbf{Post 3} & \textbf{Post 4} \\
\midrule
\textbf{TATA} \cite{tata} & $\textcolor{red}{\xmark}$ & $\textcolor{blue}{\cmark}$  & $\textcolor{red}{\xmark}$  & $\textcolor{red}{\xmark}$ \\
\textbf{MLP + RoBERTa} \cite{prakash-tayyar-madabushi-2020-incorporating} & $\textcolor{red}{\xmark}$ & $\textcolor{red}{\xmark}$  & $\textcolor{blue}{\cmark}$  & $\textcolor{red}{\xmark}$ \\
\textbf{{COLA}}~\cite{Lan_Gao_Jin_Li_2024} & $\textcolor{blue}{\cmark}$ & $\textcolor{red}{\xmark}$  & $\textcolor{blue}{\cmark}$  & $\textcolor{red}{\xmark}$ \\
\textbf{LKI-BART}~\cite{zhang-etal-2024-llm-driven} & $\textcolor{blue}{\cmark}$ & $\textcolor{blue}{\cmark}$  & $\textcolor{red}{\xmark}$  & $\textcolor{red}{\xmark}$\\
\textbf{EZSD-CP}~\cite{e26040325} & $\textcolor{red}{\xmark}$ & $\textcolor{red}{\xmark}$  & $\textcolor{blue}{\cmark}$  & $\textcolor{blue}{\cmark}$ \\
\textbf{ZSSD}~\cite{liang2022jointcl}&$\textcolor{red}{\xmark}$ & $\textcolor{red}{\xmark}$  & $\textcolor{red}{\xmark}$  & $\textcolor{blue}{\cmark}$ \\
\textbf{Dual CL} ~\cite{chen2022dual} & $\textcolor{blue}{\cmark}$ & $\textcolor{red}{\xmark}$  & $\textcolor{red}{\xmark}$  & $\textcolor{red}{\xmark}$ \\
\textbf{ZeroStance}~\cite{zhao-etal-2024-zerostance} & $\textcolor{blue}{\cmark}$ & $\textcolor{red}{\xmark}$  & $\textcolor{blue}{\cmark}$  & $\textcolor{red}{\xmark}$ \\

\textbf{{DEEM}}~\cite{wang-etal-2024-deem} & $\textcolor{blue}{\cmark}$ & $\textcolor{blue}{\cmark}$  & $\textcolor{red}{\xmark}$  & $\textcolor{blue}{\cmark}$ \\

\textbf{{LKI-BART}}~\cite{zhang-etal-2024-llm-driven} & $\textcolor{blue}{\cmark}$ & $\textcolor{blue}{\cmark}$  & $\textcolor{red}{\xmark}$  & $\textcolor{red}{\xmark}$ \\
  \midrule
\textbf{SPLAENet} & $\textcolor{blue}{\cmark}$ & $\textcolor{blue}{\cmark}$  & $\textcolor{red}{\xmark}$  & $\textcolor{blue}{\cmark}$ \\
\bottomrule
\end{tabular}}
\caption{Example posts depicting stance detection by different methods. The symbol $\textcolor{blue}{\cmark} $indicates correct predictions, while $\textcolor{red}{\xmark}$ represents incorrect predictions}
\label{Qualitative Analysis}
\end{table*}
{We observe that for the ``Support'' stance, SPLAENet successfully identifies the stance, as the source text expresses enthusiasm evident in words like ``crazy'' and positive hashtags ``capestorm'' conveying joy and trust for a weather phenomenon, and the reply confirms this excitement by validating the claim such as ``It is legit'', ``friend sent me'' maintaining the positive and trusting tone. This alignment of emotions helps to understand the sense of community, as users connect with one another's emotional expressions. Notably, in this example, SPLAENet successfully identifies the correct stance, demonstrating its capability to capture the relevant interactions between source and reply texts. Among the compared methods, only COLA \cite{Lan_Gao_Jin_Li_2024}, Dual CL \cite{chen2022dual}, ZeroStance \cite{zhao-etal-2024-zerostance}, DEEM \cite{wang-etal-2024-deem}, and LKI-BART \cite{zhang-etal-2024-llm-driven} make accurate predictions, while the remaining methods fail to infer the correct stance.
In the case of ``Query'' stance, the reply text does not convey emotions but rather poses a question using the keyword ``did.'' Although the attention mechanism can highlight phrases like ``rare natural phenomena'' and ``incredible indeed'' shows anticipation, the reply shifts to a neutral but inquisitive tone, where the interrogative ``did'' signals aniticipation. By prioritizing these emotionally charged words, SPLAENet effectively captures the emotional tone and context of the dialogue, thereby improving the overall quality of conversations. While SPLAENet performs well, only the TATA method \cite{tata}, DEEM \cite{wang-etal-2024-deem}, and LKI-BART \cite{zhang-etal-2024-llm-driven} methods correctly identify the stance, while the other methods fail in this regard.
However, in Post 3, there is a significant shift in emotional tone, as the source expresses no emotions. In the reply, the text questions the authenticity by using phrases such as ``is it'', which indicates emotional dissonance and word such as ``fake'' convey a clear negative sentiment. It can lead to misunderstanding or conflict.
While the dataset labels this as Deny based on the negative emotion, SPLAENet's current architecture appears more sensitive to the question structure than the emotional contradiction. However, the reply text is categorized as a ``Query,'' while the dataset annotation suggests it should be classified as ``Deny.''}
{In the ``Comment'' label, the hopeful question about battery life receives joy, positive and trust emotions, an affectively neutral reply (``Depends on the type''). The reply text presents a neutral statement devoid of emotional expression, allowing SPLAENet, 
along with other methods, such as EZSD-CP \cite{e26040325}, ZSSD \cite{liang2022zero}, and DEEM \cite{wang-etal-2024-deem} to identify the stance correctly.
Overall, SPLAENet showcases a superior performance by accurately classifying stances compared to existing techniques. }
\subsubsection{\textit{Statistical Analysis}}
{To rigorously evaluate the performance of SPLAENet against competing methods across the RumourEval, SemEval, and P-Stance datasets, we conducted comprehensive statistical testing. We employed the Friedman test, a non-parametric method designed to detect significant differences among multiple systems when evaluated across multiple metrics (accuracy, precision, recall, and F1-score). 
The test yielded highly significant results across all datasets showing particularly strong evidence against the null hypothesis.
The Friedman test yielded a statistic of 34.91 and a \textit{p}-value of 1.27e$-$07 in RumourEval dataset, 20.49 and a \textit{p}-value of 1.34e$-$04 in SemEval dataset, and statistic of 7.92 and a \textit{p}-value of 4.77e$-$02 in P-Stance dataset. Therefore, we reject the null hypothesis, indicating a
statistically significant difference (\textit{p} < 0.05) in performance among
the compared methods. The consistent statistical significance across multiple datasets underscores the generalizability of our approach and its potential to advance the field of stance detection.}
\section{\text{Discussion}}
\label{sec:dis}
{In this article, we propose SPLAENet, an emotion-aware dual cross-attentive neural network with
label fusion for stance detection. The model identifies the stance expressed in a reply text in relation to a source text by learning inter and intra-relationships between them through a dual cross-attention module, followed by a hierarchical attention network.
We incorporate a label fusion pipeline that analyzes the distance between contextual features and stance label information, facilitating effective mapping of features to specific stance labels.
The combination of dual cross-attention and label fusion enables the model to capture both bidirectional inter and intra-relationships between the source and reply texts, while also establishing context-dependent alignments between textual features and stance label representations.
To further enhance the discriminative power of the model, we integrate emotional alignment, particularly in contexts where emotions significantly influence opinions. This approach emphasizes important keywords and leverages emotion analysis to better understand the contextual and affective stance conveyed in the texts.
Additionally, we employ a distance-metric learning that emphasizes the contextual differences between the source and reply texts. 
To evaluate the robustness and generalizability of our approach, we conducted comprehensive experiments on three selected benchmark datasets representing distinct imbalance scenarios: (1) the highly imbalanced RumourEval dataset, (2) the moderately imbalanced SemEval dataset, and (3) the relatively balanced P-Stance dataset, as detailed in Table \ref{tab:metrics}.
This architecture proves particularly effective on the severely imbalanced RumourEval dataset, where SPLAENet achieves a 17.36\% average F1-score improvement over state-of-the-art methods.
For the moderately imbalanced  SemEval dataset, our approach maintains consistent average performance gains of 10.92\% in F1-score. 
This improved performance on skewed distributions stems from the model's dual cross-attention mechanism, which employs bidirectional interactions analysis between source and reply texts to amplify discriminative features for all classes. The label-aware fusion layer enforces a structured feature space where text representations are optimized to move closer to their correct label embedding and farther from incorrect ones.
Notably, even on the relatively balanced P-Stance dataset, the model still achieves an average gain of 11.18\% in F1-score, confirming that the architecture's benefits extend beyond imbalance handling to general stance detection performance.}

{SPLAENet achieves macro F1-scores of 51.52\% on RumourEval, 72.50\% on SemEval, and 85.58\% on P-Stance.
These variations in performance can be attributed to differences in the number of stance labels and the complexity of emotion alignment across datasets.
The number of stance labels differs across datasets: RumourEval has 4 stance labels (\textit{support, query, deny and comment}), SemEval has 3 labels (\textit{favor, against, and none}), and P-Stance has only 2 labels (\textit{favor and against}). SPLAENet captures label proximity by keeping correct labels closer to the features. 
This mechanism works best when the label space is simpler and more distinct. P-Stance with only two separable labels exhibits strong proximity. SemEval, with an added ambiguous ``none'' label, has moderate proximity. RumourEval, with fine-grained and overlapping categories like ``query'' and ``comment,'' has the weakest proximity, making it harder for the model to distinguish between them. 
P-Stance consists of politically charged content with high emotional polarization (strong positive or negative affect), which aligns well with stance labels (favor and against).
In contrast, SemEval and RumourEval feature more diverse and complex emotional expressions. Especially in RumourEval, where reply text of categories like ``query'' or ``comment'' reflect emotional dissonance, making it difficult to form a direct mapping between emotion and stance. As a result, while binary datasets like P-Stance support clearer emotion-stance relationships, the alignment between source and reply text is more complex in SemEval and RumourEval.
This approach explains the consistent improvements observed across all three datasets, establishing new state-of-the-art results and demonstrating enhanced adaptability to diverse dataset characteristics.}

\subsection{Limitations and Future Work}
{While our proposed method demonstrates notable strengths, certain limitations must be acknowledged. One key limitation is that our model currently supports only the English language. Although English is widely used, expanding the model’s capabilities to other languages is crucial for broader applicability. A multilingual approach would enable the model to analyze user-generated content across diverse linguistic and cultural contexts, making the method more inclusive and adaptable to a wider range of user-generated content.
Additionally, the model encounters challenges in processing distorted semantics resulting from incomplete or fragmented sentences, which are common in user-generated content on social media. Such informal and unstructured expressions introduce noise, potentially hindering the model’s ability to accurately capture stance.
Another area for potential improvement is the treatment of outliers in the learned representation space. While the current model reduces their effect to some extent, further refinement is needed to robustly identify and mitigate the influence of outliers. 
To address these limitations, future research can explore cross-lingual transfer learning and multilingual pretraining strategies to improve generalizability.
Furthermore, improving the model’s ability to handle these variations through enhanced pre-processing techniques and contextual understanding.
By addressing these limitations, we aim to further enhance the robustness, adaptability, and applicability of stance detection in diverse real-world contexts.}

\subsection{Theoretical and Practical Implications}
The proposed method, SPLAENet, integrates both emotion-awareness and a dual cross-attention mechanism, thereby extending the theoretical understanding of emotions and attention mechanisms to interpret stance more effectively.
By employing label fusion and distance-metric learning, SPLAENet offers a novel approach to aligning feature extraction with stance labels, which significantly improves classification performance. This approach also provides theoretical insights into how label information can be fused with attention mechanisms, facilitating the creation of more robust models for text classification tasks.
The focus of SPLAENet on emotional alignment between source and reply texts establishes a theoretical framework for understanding the role of emotions in stance differentiation. This enriches existing models surrounding the impact of emotions on the spread of misinformation and public discourse, thereby opening new avenues for exploration into the relationship between emotional content and stance.
Moreover, SPLAENet encourages the enhancement of real-time content moderation systems by identifying misleading narratives and improving misinformation detection. This capability can assist platforms and organizations in rapidly responding to and mitigating harmful content. Furthermore, social media analytics tools can benefit from SPLAENet by gaining deeper insights into public sentiment and biases, thus aiding brands and policymakers in effective online discourse analysis ~\cite{burnham2024stance,mets2024automated}. Furthermore, SPLAENet can support educational initiatives in media literacy, fostering critical thinking about how emotions and stances shape online interactions and contribute to the spread of misinformation.

In terms of computational complexity, SPLAENet comprises 383 million parameters 
and delivers a competitive inference time of 0.046 seconds.
All experiments were executed on a Linux server equipped with an Intel(R) Xeon(R) Silver 4316 CPU @2.30 GHz, 512 GB of RAM, and a 16 GB NVIDIA A16 GPU.
Regarding resource usage, SPLAENet occupies 4.4 GB of disk space when saved in the TensorFlow model format.
The model's dual cross-attention mechanism, emotion synthesis, and label-aware fusion enhance the interpretability of stance prediction. 
Overall, SPLAENet delivers strong performance while maintaining a balanced computational complexity, making it suitable for deployment in real-world scenarios.

\section{Conclusion}
In this article, we introduce a novel approach for Stance Prediction through a Label-fused dual cross-Attentive Emotion-aware neural Network called SPLAENet. As the target in the source text can be implicit, the attention mechanism plays a significant role in identifying the target. 
Our findings indicate that the attention mechanism provides more robust and contextually aware features that enhance the quality of both source and reply texts. The research unveils that such alignment of emotions between source and reply texts provides a lot of background information which is very important in analyzing stances.
The integration of label information promotes effective mapping of features to specific stance labels.
Additionally, we implement distance-metric learning between features to guarantee that semantically and contextually similar texts are closely represented in the embedding space.
Our extensive evaluation on three datasets demonstrates that the proposed method significantly outperforms baseline and state-of-the-art techniques. 
\label{con}
\\\\
\textbf{Acknowledgement}\\
We extend our gratitude to the Young Faculty Research Catalysing Grant (YFRCG) scheme, an initiative by IIT Indore, for awarding a research grant to Dr. Nagendra Kumar under Project ID: IITI/YFRCG/2023-24/03.
\bibliographystyle{elsarticle-num}

\bibliography{biblography}

\end{document}